\PassOptionsToPackage{unicode}{hyperref}
\PassOptionsToPackage{hyphens}{url}
\PassOptionsToPackage{dvipsnames,svgnames,x11names}{xcolor}
\documentclass[
  12pt]{article}

\usepackage[linesnumbered,ruled]{algorithm2e}
\usepackage{algcompatible}
\usepackage{amsfonts}
\usepackage{amsthm}
\usepackage{bm}
\usepackage{comment}

\newtheorem{theorem}{Theorem}
\newtheorem{conjecture}{Conjecture}

\newtheorem{proposition}{Proposition}

\newtheorem*{reptheorem*}{Theorem} 
\newcommand{\reptheorem}[2]{
  \begin{reptheorem*}[Restatement of Theorem~\ref{#1}]
  #2
  \end{reptheorem*}
}

\newcommand{\anon}{1}

\newcommand{\y}{\mathbf{y}}
\newcommand{\sgn}{\textrm{sgn}}
\newcommand{\z}{\mathbf{z}}

\newcommand{\x}{\mathbf{x}}
\newcommand{\X}{\mathbf{X}} 
\newcommand{\A}{\mathbf{A}} 

\newcommand{\G}[1]{\X_{#1}^\top\X_{#1}}
\newcommand{\Gi}[1]{(\G{#1})^{-1}}
\newcommand{\T}{\mathbf{T}}
\newcommand{\To}{\mathcal{T}}
\newcommand{\B}{\mathbf{B}}

\newcommand{\E}{\mathbf{E}}

\newcommand{\BOB}{\B^\top\bm{\Omega}^{-1}\B}
\newcommand{\I}{\mathbf{I}}
\newcommand{\ba}{\boldsymbol{\beta}_A}
\newcommand{\bs}{\boldsymbol{\beta}_S}
\newcommand{\bt}{\boldsymbol{\theta}}
\newcommand{\be}{\boldsymbol{\beta}}
\newcommand{\bL}{\boldsymbol{\Lambda}}
\newcommand{\bS}{\boldsymbol{\Sigma}}
\newcommand{\bep}{\boldsymbol{\epsilon}}
\newcommand{\bd}{\boldsymbol{\delta}}
\newcommand{\bz}{\mathbf{0}}

\newcommand{\hbe}{\hat{\boldsymbol{\beta}}}
\newcommand{\bet}{\boldsymbol{\eta}}
 
\newcommand{\bo}{\boldsymbol{\omega}}
\newcommand{\Oi}{\boldsymbol{\Omega}^{-1}}

\newcommand{\diag}{\textrm{diag}}

\newcommand{\W}{\mathbf{W}}
\newcommand{\sr}[2]{\frac{\left(#1\right)^2}{2#2}}
\newcommand{\argmin}[1]{\underset{#1}{\textrm{argmin}}\,\,}
\newcommand{\SI}{\mathbf{D}_{N}}
\newcommand{\TI}{\mathbf{G}_{N}}

\newcommand{\ctrp}{c_{\textrm{TRP}}}
\newcommand{\ctl}{c_{\textrm{TL}}}
\newcommand{\mineig}{\alpha_{\textrm{min}}}
\newcommand{\maxeig}{\alpha_{\textrm{max}}}

\newcommand{\nto}{\overset{n\to\infty}{\to}}

\newcommand{\lagvar}{\boldsymbol{\mu}}

\renewcommand{\lagvar}{\boldsymbol{\nu}}

\usepackage{amsmath,amssymb}
\usepackage{iftex}

\IfFileExists{upquote.sty}{\usepackage{upquote}}{}
\IfFileExists{microtype.sty}{
  \usepackage[]{microtype}
  \UseMicrotypeSet[protrusion]{basicmath} 
}{}
\makeatletter
\@ifundefined{KOMAClassName}{
  \IfFileExists{parskip.sty}{%
    \usepackage{parskip}
  }{
    \setlength{\parindent}{0pt}
    \setlength{\parskip}{6pt plus 2pt minus 1pt}}
}{
  \KOMAoptions{parskip=half}}
\makeatother
\usepackage{xcolor}
\makeatletter
\ifx\paragraph\undefined\else
  \let\oldparagraph\paragraph
  \renewcommand{\paragraph}{
    \@ifstar
      \xxxParagraphStar
      \xxxParagraphNoStar
  }
  \newcommand{\xxxParagraphStar}[1]{\oldparagraph*{#1}\mbox{}}
  \newcommand{\xxxParagraphNoStar}[1]{\oldparagraph{#1}\mbox{}}
\fi
\ifx\subparagraph\undefined\else
  \let\oldsubparagraph\subparagraph
  \renewcommand{\subparagraph}{
    \@ifstar
      \xxxSubParagraphStar
      \xxxSubParagraphNoStar
  }
  \newcommand{\xxxSubParagraphStar}[1]{\oldsubparagraph*{#1}\mbox{}}
  \newcommand{\xxxSubParagraphNoStar}[1]{\oldsubparagraph{#1}\mbox{}}
\fi
\makeatother

\usepackage{longtable,booktabs,array}
\usepackage{calc} 
\usepackage{etoolbox}
\makeatletter
\patchcmd\longtable{\par}{\if@noskipsec\mbox{}\fi\par}{}{}
\makeatother
\IfFileExists{footnotehyper.sty}{\usepackage{footnotehyper}}{\usepackage{footnote}}
\makesavenoteenv{longtable}
\usepackage{graphicx}
\makeatletter
\def\maxwidth{\ifdim\Gin@nat@width>\linewidth\linewidth\else\Gin@nat@width\fi}
\def\maxheight{\ifdim\Gin@nat@height>\textheight\textheight\else\Gin@nat@height\fi}
\makeatother
\setkeys{Gin}{width=\maxwidth,height=\maxheight,keepaspectratio}
\makeatletter
\def\fps@figure{htbp}
\makeatother

\newif\ifincludeproofs
\includeproofstrue

\makeatletter
\@ifpackageloaded{caption}{}{\usepackage{caption}}
\AtBeginDocument{%
\ifdefined\contentsname
  \renewcommand*\contentsname{Table of contents}
\else
  \newcommand\contentsname{Table of contents}
\fi
\ifdefined\listfigurename
  \renewcommand*\listfigurename{List of Figures}
\else
  \newcommand\listfigurename{List of Figures}
\fi
\ifdefined\listtablename
  \renewcommand*\listtablename{List of Tables}
\else
  \newcommand\listtablename{List of Tables}
\fi
\ifdefined\figurename
  \renewcommand*\figurename{Figure}
\else
  \newcommand\figurename{Figure}
\fi
\ifdefined\tablename
  \renewcommand*\tablename{Table}
\else
  \newcommand\tablename{Table}
\fi
}
\@ifpackageloaded{float}{}{\usepackage{float}}
\floatstyle{ruled}
\@ifundefined{c@chapter}{\newfloat{codelisting}{h}{lop}}{\newfloat{codelisting}{h}{lop}[chapter]}
\floatname{codelisting}{Listing}

\makeatother
\makeatletter
\@ifpackageloaded{caption}{}{\usepackage{caption}}
\@ifpackageloaded{subcaption}{}{\usepackage{subcaption}}
\makeatother

\ifLuaTeX
  \usepackage{selnolig}  
\fi
\usepackage[]{natbib}
\usepackage{bookmark}

\newcommand{\ourtitle}{ Formal Bayesian Transfer Learning via the \\Total Risk Prior}

\IfFileExists{xurl.sty}{\usepackage{xurl}}{} 
\hypersetup{
  pdftitle={\ourtitle},
  pdfauthor={Author 1; Author 2},
  pdfkeywords={3 to 6 keywords, that do not appear in the title},
  colorlinks=true,
  linkcolor={blue},
  filecolor={Maroon},
  citecolor={Blue},
  urlcolor={Blue},
  pdfcreator={LaTeX via pandoc}}

\begin{document}

\def\spacingset#1{\renewcommand{\baselinestretch}%
{#1}\small\normalsize} \spacingset{1}


\if1\anon
{
\title{\bf \ourtitle}
\author{Nathan Wycoff\thanks{The authors gratefully acknowledge funding from NSF/NGA Award 2428033.}\\
Department of Mathematics and Statistics \\ University of Massachusetts Amherst\\
\\
Ali Arab \\
Department of Mathematics and Statistics \\ Georgetown University
and \\
\\
Lisa O. Singh \\
Department of Computer Science \\ McCourt School of Public Policy \\ Georgetown University
}
  \maketitle
} \fi

\if0\anon
{
  \bigskip
  \bigskip
  \bigskip
  \begin{center}
    {\LARGE\bf Formal Bayesian Transfer Learning \\ via the Total Risk Prior}
\end{center}
  \medskip
} \fi

\if1\anon
\vspace{-1em}
\fi

\bigskip
\begin{abstract}
Existing methods for transfer learning struggle to deal with situations where the source datasets are limited and not guaranteed to be well-aligned with the target dataset.
A typical strategy is to use the empirical loss minimizer on the source data as a prior mean for the target parameters.
We propose using a risk minimizer conditional on source parameters instead.
This allows us to construct a joint posterior for all parameters from the source and target datasets.
As a consequence, we benefit from full Bayesian uncertainty quantification and can perform model averaging via Gibbs sampling over indicator variables governing the inclusion of each source dataset.
We show how a particular instantiation of our prior leads to a Bayesian Lasso in a transformed coordinate system and describe connections between the Maximum a Posteriori estimate associated with our approach and the Trans-Lasso method.
We find that the MAP estimator MSE-dominates the Trans-Lasso in the normal means setting when there is no regularization on the source datasets.
Finally, we perform numerical experiments finding that our approach provides superior predictive performance relative to frequentist and Bayesian baselines on a genetics application, particularly when the source data are limited.
\end{abstract}

\noindent%
\vfill

\newpage
\spacingset{1.8} 

\section{Introduction}\label{sec-intro}

The Big Data era has produced massive datasets on complex phenomena. 
\textit{Transfer learning} uses these troves, called the \textit{source datasets}, to benefit related but distinct applications with more limited data availability, called the \textit{target dataset}. 
If implemented properly, transfer learning can provide significant gains \citep{li2023targeting,hu2019statistical,hajiramezanali2018bayesian,daume2007frustratingly,wang2018deep,shang2024self}.
However, there remain important applications where the source data themselves are limited, a situation for which most existing transfer learning methods are not suitable.
This is especially so in the presence of \textit{negative transfer}: if the relationships between variables in some of the source datasets are unrelated or antithetical to the structure in the target dataset, transfer learning can be worse than simply making do with the target data.
To add to the complexity, our understanding of how to calibrate probabilistic and interval estimates in the presence of source dataset information is incomplete.
Mitigating negative transfer and calibrating probabilistic predictions are thus major open problems in the interdisciplinary field of transfer learning \citep{zhang2022survey,suder2023bayesian}, particularly in the context of moderate source data availability.
In this article, we will discuss a novel Bayesian approach to doing so and develop a methodology for performing transfer learning in linear models.

\subsection{Existing Methods and their Limitations}

We denote the target dataset by $(\X_0, \y_0)$, and assume that $\y_0 \sim F_{\be_0}(\X_0)$, where $\be_0$ are the parameters of the target distribution.
We use similar notation for the source datasets: $\y_k \sim F_{\be_k}(\X_k)$ for all $k\in\{1, \ldots, K\}$.
Note that each source dataset has its own possibly distinct parameter.
We assume that all datasets share the same variables and thus, each $\X_k$ has the same column dimension $P$.

Though transfer learning has a long history in computer science and related fields (see e.g. the surveys \citet{pan2009survey,zhuang2020comprehensive,weiss2016survey}), there has been a renaissance in the statistics literature recently where fascinating statistical properties have been discovered with considerable practical consequence.
\citet{li2022transfer} showed that transfer learning can improve rates in the minimax setting for linear models with sparse coefficients, and interesting frequentist results have also been presented for, among other settings, Generalized Linear Models \citep{tian2023transfer,li2024estimation}, quantile regression \citep{tian2023transfer} and graphical models \citep{li2023transfer}.
The approach of \citet{li2022transfer}, called \texttt{Trans-Lasso}, may be summarized as performing two Lasso regressions, first, one on all relevant source datasets concatenated, and then one on the residuals of the target dataset given that first regression's parameters. 
The sum of these two coefficients is used as an estimate for the target coefficient.
\cite{abba2024bayesian} showed that there can be advantages in using a Bayesian approach instead of that second Lasso.
These methods all fall within a common high level strategy for transfer learning, which proceeds by first building an estimator $\tilde\be$ using the source data, and then, when fitting the target data, controlling the difference between $\be$ and $\tilde\be$ using either regularization or a prior.

Most prior work describable as ``Bayesian transfer learning" falls into two categories: classical Bayesian hierarchical models and two-stage approaches similar to those just discussed.

\subsubsection{Bayesian Hierarchical Models and Transfer Learning}
First, take Bayesian hierarchical models. 
These are formal Bayesian methods which allow information sharing across datasets. 
Hence, some authors, such as \citet{suder2023bayesian}, categorize them as Bayesian transfer learning. 
However, there is a certain directionality in transfer learning which is not captured by the hierarchical model: we want to transfer information \textit{from} the source datasets \textit{to} the target dataset.
Hierarchical models, by contrast, are involved in all-to-all, omnidirectional information transfer, and other authors \citep[such as][]{bull2023hierarchical} have thus categorized them as multi-task learning \citep{zhang2021survey}.

We know of two important works which attempt to inject directionality into a hierachical model.
First, \citet{zhang2024covariate}: who privilege the target dataset such that this directionality is actually reversed: 
the target parameters are used as prior location parameters for the \textit{source} parameters, the reverse of the approach taken by e.g. \citet{abba2024bayesian}.
Though it betrays our conceptual intuition, the math works out, and this intriguing approach allows for Bayesian transfer learning which avoids negative transfer by doing Gibbs sampling on parameter-specific dataset indicators.
Second, the BLAST approach of \citet{jamshidian2025bayesian}.
This involves assuming that there are two underlying regression coefficients: one associated with the target dataset and any relevant source datasets, and one shared by the other source datasets. 
Bayesian inference is done on which source dataset goes in which pile. 
Target-source asymmetry is introduced by forcing membership of the target dataset in one group (whereas other datasets have indicator variables giving group membership which are averaged over during MCMC).
However, this approach imposes significant homogeneity assumptions: datasets which are not relevant to the target may nonetheless be included in its group because they are even less related to the other irrelevant sources.

\begin{figure}
    \centering
      \begin{minipage}[b]{0.7\textwidth}
        \includegraphics[width=\linewidth]{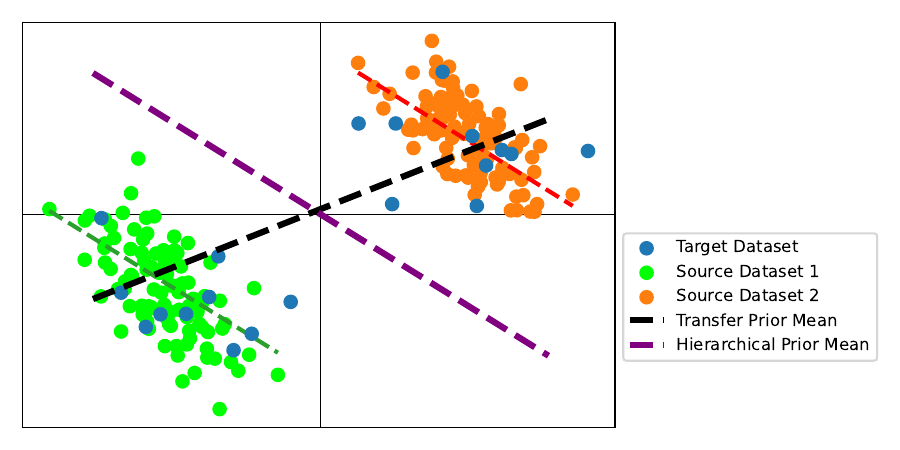}
    \end{minipage}%
    \hfill
    \begin{minipage}[b]{0.29\textwidth}
      \centering
      \hspace{-2em}
      \raisebox{0.25\height}{
      \includegraphics[width=\linewidth]{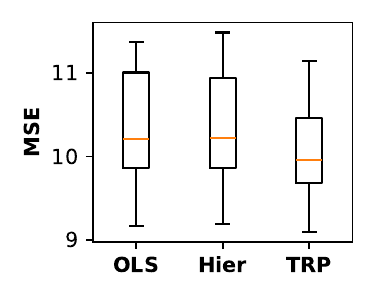}
      }
    \end{minipage}
    \caption{
    \textit{Left:} Toy dataset, an instantiation of Simpson's Paradox, with prior data information available via our proposed prior but not a hierarchical prior.
    \textit{Right:} Boxplot giving out-of-sample MSE for OLS without prior information, a hierarchical model, and a transfer model, repeated 10 times with different random numbers.
    }
    \label{fig:simpsons}
\end{figure}

Despite these special considerations, these approaches are affected by an overarching difficulty.
Loosely speaking, hierarchical models share information by shifting estimates of a given parameter towards the overall average of parameters, which is qualitatively different than frequentist transfer learning methods like \citet{li2022transfer} and descendants, which fit a model to all source datasets jointly.
This difference is best illustrated by example.
Figure \ref{fig:simpsons} presents a pathological dataset which is an instantiation of Simpson's paradox.
The Left subfigure shows two generated source datasets (green and orange points) within the setting of univariate linear regression with non-overlapping set of $x$ variables. 
Within each dataset, the relationship between the predictor and the response is decreasing. 
However, the overall average values for each dataset are such that if we were to fit them jointly, we would observe a positive relationship. 
We subsequently consider a target dataset (blue points) which consists of observations from within both of these clusters. 
The sign of the regression coefficient for the target dataset is thus the opposite of those for both source datasets considered individually. 
Hence, a Bayesian hierarchical model, which averages the source dataset coefficients to form a prior mean, is not helpful (see Figure \ref{fig:simpsons}, Left, the dashed purple line's relationship to the target dataset). 

In this article, we will propose the \textit{Total Risk Prior} (TRP), which would in this case use a prior mean given by the minimizer of risk summed over both source datasets, shown as the dashed black line in the figure.
This results in a useful prior which improves predictive performance (see right panel).
Of course, on this example, the careful analyst who visualizes their datasets before modeling could have noticed this and combined the two sources.
However, our aim in this article is to develop a method which would automatically accommodate this structure, which is essential in high dimensional, complex settings.

\subsubsection{The Incumbent 2-Stage Transfer Learning Paradigm}

The second high level category of Bayesian transfer learning consists of two-stage methods involving the use of hyperparameters in the target parameters' prior derived from the source datasets.
\citet{raina2006constructing} and \citet{abba2024bayesian} fit into this group, learning scale and location parameters, respectively.
This can be an advantage in that the analyst does not need access to the source datasets themselves, which can mean reduced computation and privacy risk.
There are also important downsides, however, as these two-stage methods cannot retroactively remove source datasets causing negative transfer, and they typically ignore uncertainty in the source dataset estimates.
These methods are, to quote \citet{pramanik2025modeling} discussing their own method, ``post-hoc'': the decision to perform transfer learning is made outside of the Bayesian formalism, which enters into play after the source datasets have been exploited.
By \textit{formal Bayesian transfer learning}, we will mean the specification of a joint prior $P(\be_0, \be_1, \ldots, \be_K)$ which is then simply combined with the likelihoods $\prod_{k=0}^K \mathcal{L}(\y_k;\be_k,\X_k)$ via Bayes' rule to yield a marginal posterior $P\left(\be_0|(\X_0, \y_0),\ldots,(\X_K,\y_K)\right)$ after integrating over $\be_1,\ldots,\be_K$.
By contrast, the two-stage methods we are presently discussing instead specify a prior only on $\be_0$ of the form $P\left(\be_0 | (\X_1,\y_1), \ldots, (\X_K,\y_K)\right)$, or alternatively $P(\be_0|\hat\be_1,\ldots,\hat\be_K)$ for some source dataset-derived estimates $\hat\be_1,\ldots,\hat\be_K$, and treat these hyperparameters as fixed.
Practical consequences of this include not accounting for uncertainty on source parameters and an unclear path towards using Bayesian means to select which source datasets to include.
\citet{li2022transfer} propose Q-aggregation to mitigate negative transfer, but 1) this cannot accommodate computationally intensive methods, like \citet{abba2024bayesian}, and 2) while it has mathematical guarantees against including pernicious source datasets, it may miss beneficial ones.
TRADER \citep{lai2026bayesian} is a 2-stage method that bucks the trend, and is able to select from a subset of source estimators to input into a weighted average within an MCMC algorithm.
However, like the other 2-stage methods it considers only point estimates from the source datasets, rather than the full source datasets themselves, which ignores uncertainty that can be important when the source datasets are not plentiful.

Of course, these two categories are not exhaustive of the Bayesian transfer learning literature: there have been, for instance, interesting methods which avoid estimating target parameters altogether, instead operating directly on source model predictions \citep[e.g.][]{hickey2024transfer,pramanik2025modeling,presicce2026bayesian}.
However, in this article, we are interested in estimating target parameters, where the proposed dichotomy for the most part holds.

\subsection{Our Contributions}

Thus, the current state of affairs is that one must choose either 1) A formal Bayesian model, or 2) the qualitative behavior of frequentist two-stage transfer learning, but not both. The purpose of this article is to dissolve this dilemma.

To this end, we propose using a prior distribution conditional on the values of source dataset \textit{parameters} for the target parameter, as opposed to the source data themselves, as is done by Bayesian two-stage approaches.
But unlike existing hierarchical approaches, our prior encodes the belief which might be held if we thought that minimax frequentist transfer learning methods would be appropriate if only we had bigger datasets.
The main idea, described in Section \ref{sec:trp}, is to specify that $\be_0$ is close to the minimum, not of the observed penalized sum of squares given the source data, but rather the \textit{expectation} of this quantity under a particular setting of $\be_1, \ldots, \be_K$. 
In the language of Bayesian decision theory, we propose to use as a prior hyperparameter for $\be_0$ not the minimizer of total empirical loss on source datasets, but rather the minimizer of total risk.
This gives us the desired qualitative behavior while being free of any data.
We can then jointly perform Bayesian inference over all source datasets and the target dataset simultaneously, specifying indicator variables that control whether a given source dataset is used for transfer learning or left to be fit independently. 
We then present a source-selection consistency result for these indicators insofar as if the target coefficients are exactly representable as a precision-weighted combination of source coefficients, that combination is selected with probability 1 in the limit.
In the more realistic case where there is no such ``exact transfer", we find that the indicator posterior places mass on all configurations with probability approximately inversely proportional to the difference between anticipated and actual target coefficients.

In Section \ref{sec:map}, we study the MAP estimate associated with this model, which may be solved via a standard Lasso call on an augmented dataset.
We find that if we stop a particular optimization algorithm for the MAP after one iteration, we recover the Trans-Lasso of \citet{li2022transfer}.
We then turn to the normal means setting, and develop closed form expressions for the MAP using convex analysis.
Also in the normal means setting, we find that if there is no regularization used for fitting the source dataset, our MAP estimator dominates the Trans-Lasso under $\ell_2$ loss with known error variance.

Section \ref{sec:app} presents numerical experiments on a genetic application, where our method is compared to the Trans-Lasso and BLAST as well as classical baselines.
We find that this joint inference over all parameters leads to greater predictive accuracy than existing methods at the cost of greater computational expense.
It is most suitable for situations where not only target data but even source data are scarce, and it is important to squeeze maximal information out of all datasets, particularly when it is not clear that all source datasets will be informative and negative transfer is to be avoided. 

These numerical experiments are conducted via the new open source \texttt{Python} package \texttt{bayes\_trans}\footnote{
\if1\anon
\url{https://github.com/NathanWycoff/bayes_trans}
\fi
\if0\anon
\url{www.github.com/anonymized/anonymized}; [please see supplementary material for python package during review process.]
\fi
}
based on \texttt{JAX} \citep{jax2018github} which enables efficient multicore computing on CPUs and various accelerators such as GPUs.

\subsection{Problem Setting and Notation}

We denote the number of observations in dataset $k$ by $N_k$, the total sample size by $N=\sum_{k=0}^K N_k$ , and for each dataset $\y_k = \X_k\be_k + \bep_k$ with $\bep_k$ a column vector containing iid mean-zero Gaussian noise with common finite $\sigma^2$.
If $\be_A = [\be_0^\top, \ldots, \be_K^\top]^\top$ is a vector containing all dataset parameters concatenated together, let $\E_k \in \mathbb{R}^{P\times (K+1)P}$ be the matrix extracting the regression coefficients for the $k$th dataset, i.e. it is a block matrix with a $P\times P$ identity in the $k$th block and zeros elsewhere.
For a generic matrix $\A$, $\A^\top$ denotes its transpose and $\A^{-\top}$ the inverse of its transpose.
For a matrix $\A$ with real eigenvalues, $\mineig(\A)$ and $\maxeig(\A)$ denote its  minimum and maximum eigenvalues, respectively.
When we write $\x\sim N\left(\boldsymbol\mu, \boldsymbol\bS\right)$, we mean $\x$ is multivariate normal with mean $\boldsymbol\mu$ and covariance $\boldsymbol{\bS}$.

\section{Bayesian Transfer via the Total Risk Prior}
\label{sec:trp}

Assume for now that all source datasets are informative.
The central conundrum is then the following. 
We'd like to stay close to the approaches of \citet{li2022transfer,abba2024bayesian} and others in first computing the loss minimizer on the source datasets:
\begin{align}
    \tilde\be \gets \underset{\be\in\mathbb{R}^P}{\textrm{argmin}} \,
    \frac1{2}
    \sum_{k=1}^K
    \Vert \y_k - \X_k \be \Vert_2^2   
    + \gamma(\be)
    \,,
\end{align}
where $\gamma$ is some penalty function, and then using $\tilde\be$ as a prior location parameter for $\be_0$.
However, $\tilde\be$ is a function of the source data and can't be used in the joint prior $P(\be_0, \ldots, \be_K)$ in the formal Bayesian setting.
Our key conceptual contribution is to retain the functional form of the minimization while taking its expectation conditional on a particular setting for the source parameters:
\begin{equation} \label{eq:To}
    \underset{\be\in\mathbb{R}^P}{\textrm{argmin}} \,
    \frac1{2}
    \sum_{k=1}^K
    \underset{\y_k\sim F_{\be_k}(\X_k)}{\mathbb{E}}
    \left[
    \Vert \y_k - \X_k \be \Vert_2^2   
    \right]
    +\gamma(\be)
    \,.
\end{equation}
In other words, instead of minimizing empirical loss, we propose to minimize risk.
We then define $\To$ as the $\mathbb{R}^{KP}\to\mathbb{R}^P$ map which takes some candidate regression coefficient setting $\bs := [\be_1^\top\,\ldots\,\be_K^\top]^\top\in\mathbb{R}^{KP}$ across all source datasets to Expression \ref{eq:To}, the regularized expected squared error minimizer under that coefficient setting, denoted $\To(\be_S)$.
We refer to $\To$ as the \textit{transfer operator}.
It is defined such that whatever the coefficients $\bs$ may be, $\To(\bs)$ will be well-suited to make predictions on the entirety of the source datasets in the $\ell_2$ predictive sense.
We can use the transfer operator to develop a prior distribution jointly on $\ba:=[\be_0^\top \, \be_1^\top \, \ldots \, \be_K^\top]^\top\in\mathbb{R}^{(K+1)P}$ by taking:
\begin{equation}\label{eq:transprior}
    P(\be_0|\be_1,\ldots,\be_K) 
    \propto
    d\left(\Vert \be_0 - \To(\be_1,\ldots,\be_K)\Vert\right)\,.
\end{equation}
Here, $d$ is some suitable nonnegative decreasing function and $\Vert\cdot\Vert$ is some norm or other measure of size.
We term a prior formed via this strategy as being a \textit{Total Risk Prior} (TRP) as it involves the argument minimizing the sum of risk over all source datasets conditional on a particular setting of source data parameters.

Under our assumptions, the expectation in \ref{eq:To} evaluates to
\begin{align}
    & 
    \mathcal{T}(\bs) = 
    \underset{\be\in\mathbb{R}^P}{\textrm{argmin}} \,
    \frac1{2}
    \sum_{k=1}^K
    (\be_k-\be)^\top\X_k^\top\X_k(\be_k-\be)
    + \gamma(\be)
    \,.
\end{align}
From the perspective of an optimization algorithm, this expression is very similar to a least-squares regression problem regularized by $\gamma$, and similar computational approaches may be used for solving both problems, as we will establish.
In addition, this similarity guides us as to when we might expect the 
optimization will have a unique minimizer, meaning that $\To_{\bet}$ is a well-defined (i.e. single-valued) function.
This is most simply obtained if $\gamma$ is strongly convex, where its pointwise sum with the quadratic form will also be strongly convex and thus uniquely minimized even if the quadratic form is only positive-semidefinite.

Bayesian computation may be conducted in the general case via Markov chain Monte Carlo \citep[MCMC;][]{brooks2011handbook} methods such as Metropolis within Gibbs \citep{metropolis1953equation,geman1984stochastic,gelfand1990sampling,tierney1994markov} which require only unnormalized density evaluations by 1) solving the stochastic program in Expression \ref{eq:To} then 2) plugging the result into Equation \ref{eq:transprior}.
This is because the mapping
\begin{align*}
    \begin{bmatrix}
        \be_0\\ \be_S
    \end{bmatrix}
    \to
    \begin{bmatrix}
        \be_0\\\To\be_S
    \end{bmatrix}
    \quad 
    \textrm{is a shear, and for smooth $\To$, has Jacobian given by}
    \quad
    \begin{bmatrix}
        \I & \nabla \To\be_S \\ \bz & \I
    \end{bmatrix}\,,
\end{align*}
such that it does not contribute to the normalizing constant regardless of $\To$.
This allows us to avoid the doubly-intractable inference paradigm that sometimes accompanies distributions with embedded optimization problems.
However, we will not proceed at this level of generality in this article, as we will find that making additional assumptions will be quite profitable in terms of computational savings.

We make one more addition to $\To$ before continuing.
To mitigate negative transfer, we will define a binary vector $\bet\in\{0,1\}^K$, where $\eta_{k}=1$ indicates that dataset $k$ is to be included in transfer, and $\eta_k=0$ that it should not be.
We then add these to the definition of $\To$:
\begin{align}
    \To_{\bet}(\be_S)
    :=    \,
    \underset{\be\in\mathbb{R}^P}{\textrm{argmin}} \,
    \frac1{2}
    \sum_{k=1}^K
    \eta_k
    (\be_k-\be)^\top\X_k^\top\X_k(\be_k-\be)
    + \gamma(\be)
    \,.
\end{align}
We will thus think of the transfer operator as being parameterized by $\bet$, denoting it as $\mathcal{T}_{\bet}$ when this is to be emphasized.
Inference on $\bet$ can be conducted via Gibbs sampling, inheriting the predictive benefits of Bayesian model averaging from the tradition of performing Gibbs sampling for variable selection \citep{george1993variable} while avoiding the technical difficulties of changing measure dimension \citep{green1995reversible}.

Computability of $\To(\be_S)$ for given $\be_S$ depends on the complexity of $\gamma$. 
For instance, if we were to follow \citet{li2022transfer} closely, we would use $\gamma(\be) = \tau \Vert \be \Vert_1$. 
In this case, we are left with an $\ell_1$ regularized quadratic cost, which is similar to the Lasso optimization. 
This could therefore be resolved via either a quadratic program or a coordinate descent method, allowing for a relatively efficient Metropolis within Gibbs Algorithm.
This approach makes posteriors based on the TRP computable in principle for a wide range of penalizers $\gamma$.
However, we will in this article instead primarily focus on the case where $\gamma(\be) = \frac{\tau}{2} \Vert\be\Vert_2^2$.
This is because the transfer operator is then linear and has a closed form:
\begin{align}
    \mathcal{T}_{\bet}(\bs) = \left(
    \sum_{k=1}^K\eta_k \X_k^\top\X_k+\tau\I
    \right)^{-1}
    \sum_{k=1}^K\eta_k \X_k^\top\X_k \be_k 
    := \T_{\bet} \bs
    \,,
\end{align}
where $\T_{\bet}\in\mathbb{R}^{P\times KP}$, which we'll call the \textit{transfer matrix}, is given by:
\begin{equation}
    \T_{\bet} = 
    \left(
    \sum_{k=1}^K\eta_k \X_k^\top\X_k+\tau\I
    \right)^{-1}
    \begin{bmatrix}
        \eta_1\X_1^\top\X_1 & \ldots & \eta_K\X_K^\top\X_K
    \end{bmatrix}
    \,,
\end{equation}
and is the matrix representation of $\mathcal{T}_{\bet}$.
If $\tau = 0$, this is a precision-weighted average of each regression coefficient; see Appendix \ref{ap:tmath} for other basic properties of this matrix.
In any event, using the $\ell_2$ regularizer leads to the following TRP:
\begin{equation}
    P(\be_0|\be_1,\ldots,\be_K,\bet,\lambda_t,\tau)\propto
    d\left(\lambda_t 
    \left\Vert  \left(\E_0 - 
    \begin{bmatrix}
        \bz & \T
    \end{bmatrix}
    \right)\be_A
    \right\Vert\right) 
    := d\left(\lambda_t\Vert\B\ba\Vert\right) 
    \,,
\end{equation}
where we have defined $\B\in\mathbb{R}^{P\times (K+1)P}$ to make clear that this may be understood as a prior in the $P$-dimensional space spanned by $\B$'s rows.
We will give special attention to two choices of $d$ and $\Vert\cdot\Vert$. 
Choosing $d(t) = \exp(-\frac1{2\sigma^2}t^2)$ and the $\ell_2$ norm yields a Gaussian TRP, while choosing $d(t) = \exp(-\frac1{\sigma}t)$ and the $\ell_1$ norm yields a Laplace TRP.
Division by $\sigma^2$ or $\sigma$ here is critical to maintaining unimodality of the posterior \citep{park2008bayesian}.
Each of these priors may be viewed as specifying prior independence in the space $\z := \B\be_A\in\mathbb{R}^{P}$.

\subsection{Model Specification and Gibbs Sampling}

We now illustrate how the TRP may be used in practice by specifying a complete statistical model to be used in our numerical experiments.
We then describe the conditional posterior distributions this induces, leading to a Gibbs sampler.
In the case of a Gaussian TRP, we of course have a Gaussian conditional posterior for $\be_A$:
\begin{align}
    \label{eq:ridge_post}
    P(\be_A|\lambda_t,\tau,\bet,\bL_0,\sigma^2) = 
    N
   \left(
   \bS_{\be} \X_A^\top\y_A
   ,
   \sigma^2\bS_{\be}
   \right)
   \,,
\end{align}
where we have assumed a Gaussian prior $\be_S|\sigma^2\sim N(\mathbf{0}, \sigma^2\bL_0^{-1})$ on source coefficients. 
\begin{align}
    \label{eq:bigmats}
    \bS_{\be} = 
   \left(
    \X_A^\top\X_A + \lambda_t^2\B^\top\B + \bL_0 
   \right)^{-1} ;
    & & \X_A = 
    \begin{bmatrix}
        \X_0 & \bz & \ldots & \bz \\ 
        \bz & \X_1 & \ldots & \bz \\ 
        \vdots & \vdots & \ddots & \bz\\
         \bz & \bz & \ldots & \X_K
    \end{bmatrix};
    &&
    \y_A = \begin{bmatrix}
        \y_0 \\
        \y_1 \\  
        \vdots \\ 
        \y_K
    \end{bmatrix}
     \,,
\end{align}
where
$\X_A\in \mathbb{R}^{\left(\sum_{k=0}^K N_k\right) \times (K+1)P}$, 
$\y_A \in\mathbb{R}^{\sum_{k=0}^K N_k}$ and 
$\bS_{\be}\in\mathbb{R}^{(K+1)P\times (K+1)P}$.
We simply use $\bL_0 = \lambda_p^2\I$ in this article. 
Since $\bS_{\be}$ is of dimension $(K+1)P\times (K+1)P$, a standard Cholesky-based sampling approach has computational cost of order $K^3P^3$, which is intractable for our application case study where $K=36$ and $P=399$.
Following \citet{bhattacharya2016fast}, we exploit the block-diagonal + low rank structure for sampling with $O(KP^3)$ computation.

For a Laplace prior, we can exploit literature on the Bayesian Lasso to perform Gibbs sampling via an auxiliary variable scheme \citep{carlin1991inference,park2008bayesian,hans2009bayesian} using the representation of the Laplace distribution as an exponential scale mixture of Gaussians \citep{andrews1974scale}:
\begin{align}
    \exp  \left(
    -\frac{\lambda_t}{\sigma}
    \left\Vert
    \B \be_A 
    \right\Vert_1
    \right)
    \propto
    \int_{\mathbb{R}_+^P}
    \left(
    \frac{e^{-\frac1{2\sigma^2} \sum_{p=1}^P\frac{(\B\be_A)_p^2}{\omega_p}}}{\sqrt{\prod_{p=1}^P \sigma^2\omega_p}}
    \right)
    \left(
    \lambda_t^{2P}
    e^{-\frac{\lambda_t^2}{2} \sum_{p=1}^P \omega_p}
    \right)
    d\bo \,.
\end{align}
We to this end introduce into our model the auxiliary variables $\bo\in\mathbb{R}^P_+$ with prior conditional on $\lambda_t$ given by an exponential distribution with rate $\frac{\lambda^2_t}{2}$, so that $(\B\be_A)_p|\omega_p,\sigma^2\sim N(0,\sigma^2\omega_p)$.
This leads to a similar posterior conditional $P(\be_A|\bo, \tau, \bet, \bL_0,\sigma^2)$ as in Equation \ref{eq:ridge_post}, but now with $\bS_{\be} = \left(\X_A^\top\X_A + \BOB + \bL_0\right)^{-1}$, where $\boldsymbol{\Omega} = \diag(\bo)$.

Of course, alternative priors for $\bo$ may be considered.
The Gaussian model of Equation \ref{eq:ridge_post} may be viewed as a special case of this extended model with prior $P(\bo)=\delta_{\lambda_t^{-2}}$, a point mass at $\lambda_t^{-2}$ in each coordinate. 
We could obtain the horseshoe prior \citep{carvalho2010horseshoe} on $\B\be_A$ by placing an inverse gamma prior on $\bo$ instead.

We place the standard prior $P(\sigma^2)\propto \frac{1}{\sigma^2}$ on the error variance, which leads to the standard Inverse Gamma posterior full conditional on $\sigma^2$.
In particular, we have:
\begin{align}
    P(\sigma^2 | \be_A, \lambda_t, \lambda_p, \bo, \bet) 
    =
    \Gamma^{-1}
    \left(
    \frac{N+(K+1)P}{2}
    ,
    \frac{
    \Vert\y_A - \X_A\be_A\Vert_2^2
    +
    \Vert\boldsymbol{\Omega}^{-\frac1{2}}\B\be_A\Vert_2^2 
    + 
    \lambda_p^2
    \Vert
    \be_S
    \Vert_2^2
    }{2}
    \right)
    \,,
\end{align}
where the second argument gives the rate of the associated Gamma distribution. 

We place Half-Cauchy priors on $\lambda_t$ and $\lambda_p$ with unit scale.
Recall the representation of Half-Cauchy distributions as an inverse gamma hierarchy \citep{makalic2015simple}, where:
\begin{align}
    x \sim \textrm{HalfCauchy(0,A)}
    \iff 
    a \sim \Gamma^{-1}\left(\frac1{2}, \frac{1}{A^2}\right); 
    x^2 |a \sim \Gamma^{-1}\left(\frac1{2}, \frac{1}{a}\right) 
    \,.
\end{align}
This hierarchy is designed for $x$ a scale parameter, so that $x^2$ is a variance and enters its likelihood in inverse-gamma form.
Both $\lambda_t$ and $\lambda_p$ are instead rate parameters: $\lambda_t^2$ is the rate of the exponential prior on $\bo$, and $\lambda_p^2$ is the prior precision of $\be_S$, so each enters its likelihood in gamma form.
We therefore apply the hierarchy to the reciprocals $x_t := 1/\lambda_t$ and $x_p := 1/\lambda_p$, which are the corresponding scales.
Since the standard Half-Cauchy distribution is invariant under $x\mapsto1/x$, this leaves the priors on $\lambda_t$ and $\lambda_p$ themselves unchanged, while an inverse gamma prior on $x_t^2 = \lambda_t^{-2}$ is a gamma prior on $\lambda_t^2$, which is conjugate.
Concretely, we introduce two auxiliary variables $a_t,a_p$ with Inverse Gamma priors of shape $\frac1{2}$ and scale $1$, and take $\lambda_t^2|a_t \sim \Gamma\left(\frac12, \frac{1}{a_t}\right)$ and $\lambda_p^2|a_p\sim\Gamma\left(\frac12,\frac{1}{a_p}\right)$, where here and below the second argument of $\Gamma(\cdot,\cdot)$ is a rate.
With the iid exponential prior of rate $\frac{\lambda_t^2}{2}$ on $\bo$ described above, the full conditional for $\lambda_t^2$ is then:
\begin{align}
    P(\lambda_t^2|a_t,\bo)
    =
    \Gamma\left(
    P+\frac{1}{2},
    \frac1{a_t}
    +
    \frac1{2}\sum_{p=1}^P \omega_p
    \right)
    \,,
\end{align}
and similarly:
\begin{align}
    P(\lambda_p^2|a_p,\be_S,\sigma^2) = 
    \Gamma
    \left(
    \frac{KP+1}{2},
    \frac1{a_p}
    +
    \frac{\Vert\be_S\Vert_2^2}{2\sigma^2} 
    \right)
    \,.
\end{align}
The auxiliary variables have full conditionals given by 
$a_t\sim \Gamma^{-1}\left(1,1+\lambda_t^2\right)$
and
$a_p\sim \Gamma^{-1}\left(1,1+\lambda_p^2\right)$.

The $\bo$ updates are as in the Bayesian lasso, except that we consider the elements of $\B\be_A$ rather than the regression coefficients directly.
In particular,
\begin{align}
    P\left(\frac1{\omega_p}\Bigg|\be_A,\sigma^2,\lambda_t\right) = 
    \mathcal{W}
    \left(
    \sigma\frac{\lambda_t}{|(\B\be_A)_p|}
    ,
    \lambda_t^2
    \right)
    \,,
\end{align}
where this is a Wald distribution with mean $\mu$ and scale $s$ whose density is:
\begin{align}
    f(x;\mu,s) = 
    \sqrt{\frac{s}{2x^3}}
    e^{\frac{-s(x-\mu)^2}{2\mu^2x}}
    \,.
\end{align}


The update $\bet$ is done one entry at a time.
For a given $k$, we thus compute the relevant posterior conditional:
\begin{align}
    \exp\left(
    -\frac1{2\sigma^2}
    \be_A^\top
    \B_{\bet^*}^\top\Oi\B_{\bet^*} 
    \be_A
    \right)
    \times 
    \rho^{\eta_k^*}
    (1-\rho)^{1-\eta_k^*}
\end{align}
for $\eta^*_k = 0$ and subsequently for $\eta^*_k = 1$ and sample with probability proportional to those expressions.
Conditional on $\bet$, $\rho$ of course has a Beta distribution with parameters $\sum_{k=1}^K \eta_k+0.5, K-\sum_{k=1}^K\eta_k + 0.5$.

We found in practice that the conditional distribution for $\bet$ as a whole was multimodal with significant energy barriers preventing this simple algorithm from sufficiently exploring probable states.
It is possible to conduct ``Rao-Blackwellized''\citep{robert2021rao} updates for $\eta_k$, with either $\bo$ or $\be_A,\sigma^2$ integrated out.
In particular, integrating out $\be_A$ leads to heavier tails, but these were insufficient to achieve reasonable mixing.
We instead found parallel tempering \citep{swendsen1986replica,geyer1991computing} to be critical.

We also place a Half-Cauchy on $\tau$. 
However, because of its position as a coefficient for a matrix sum under a matrix inverse and determinant, we do not expect that a closed-form Gibbs update is possible.\footnote{Indeed this expression is similar to that of a Gaussian process's nugget, for which we are aware of no closed-form Gibbs updates.}
We simply conduct a random walk Metropolis step with Gaussian proposal on this parameter, using the standard procedure to set the proposal variance \citep{robbins1951stochastic} and rejecting any nonpositive proposals.


\subsection{Source Selection Consistency}
\label{sec:theory}

We now investigate the behavior of the posterior in large samples.
Though our study is focused on the small data setting, we will find that there is still important intuition to be gained from an asymptotic perspective.
We use a bar to denote a limiting quantity. 
In the case of identified parameters, such as $\be$ under our further assumptions, this also gives the  data-generating value.
Assume that $\frac1{N_k}\G{k}\nto\bS_{k} \succ \mathbf{0}$ and that $\frac{N_k}{N}\to\pi_k\in(0,1)$ for each $k\in\{0,\ldots,K\}$.
We have that:
\begin{align}
    \T\overset{n\to\infty}{\to} 
    \left(
    \sum_{k=1}^K 
    \pi_k\bS_k
    \right)^{-1}
    \begin{bmatrix}
        \pi_1\bS_1
        & \ldots & 
        \pi_K\bS_K
    \end{bmatrix} 
    := \bar\T\,.
\end{align}
Note that this quantity does not depend on $\tau$, which is asymptotically inert, so we will not discuss it further in this subsection. 
However, it is nevertheless critical in practical situations with small source datasets to avoid ill-posedness. 
Since the prior covariance matrix is convergent, classical Bernstein-von-Mises results hold, and $\be,\sigma^2$ converge simply to the data-generating values $\bar\be,\bar\sigma^2$.
More interesting is the asymptotic model selection behavior, which we now study by determining the limiting distribution for $\bet$.

\begin{theorem}
    Under the aforementioned assumptions, if there is a \emph{unique} setting of the source data indicators $\bet^*$ such that $\bar\T_{\bet^*}\bar\be_S = \bar\be_0$, then:
    \begin{align}
    \lim_{N\to\infty}
    P(\bet=\bet^*|\y_A,\X_A) = 1 
    \iff 
    \mathbb{E}_{\lambda_t}\!\left[\lambda_t^{P}\right] = \infty
    \,.
    \end{align}
    
\end{theorem}
\begin{proof}
    See Appendix \ref{ap:postasymp}.
\end{proof}
This shows that our approach is consistent in estimating the inclusion indicators under the assumption that there is an exact match between the target and some combination of sources, so long as the prior distribution on the penalty parameter is sufficiently heavy-tailed. 
This is reminiscent of the heavy-tailedness condition required for consistency in the context of $g$ priors for Bayesian model selection \citep{zellner1986assessing,liang2008mixtures}.
This is of course satisfied by our Half-Cauchy prior on $\lambda_t$ since all moments $P\geq 1$ are divergent.

If we are not in this exact transfer setting, $P(\bet|\y_A,\X_A)$ instead asymptotically weights the sources according to its prior times the following expression (see again Appendix \ref{ap:postasymp}):
\begin{align*}
\int_0^\infty 
    \lambda_t^{P}
    e^{-\frac{\lambda_t}{\bar\sigma}\Vert\bar\be_0 - \bar\T\bar\be_S\Vert}
    P(\lambda_t)
    d\lambda_t \,,
\end{align*}
which is nonincreasing in $\Vert\bar\be_0 - \bar\T\bar\be_S\Vert$ regardless of $P(\lambda_t)$ and $\Vert\cdot\Vert$. 

For example, under a Laplace prior on $\be_A$ and a Half Cauchy prior on $\lambda_t$, we obtain (ibid):
\begin{align}
\lim_{N\to\infty}
P(\bet|\y_A,\X_A) \propto 
\frac{P(\bet)}{\Vert\bar\be_0 - \bar\T\bar\be_S\Vert_1^{P+1}}
\left(
1+
O
\left(
\Vert\bar\be_0 - \bar\T\bar\be_S\Vert_1^{-2}
\right)
\right)
\,.
\end{align}
This means that the Bayes factor comparing two different source settings may be approximated by the ratio of the residual term $\bar\be_0 - \bar\T_{\bet} \bar\be_S$ norms for the two candidates, where this approximation holds when the norm of the deviations are large.

As an example, if there is a single source ($K=1$), then $\bar\T_{\eta=1}=\I$ and $\bar\T_{\eta=0}=\mathbf{0}$, so that the posterior odds in favor of inclusion are
\begin{equation}
    \lim_{N\to\infty} \frac{P(\eta=1|\y_A,\X_A)}{P(\eta=0|\y_A,\X_A)}
    \overset{\be_A \textrm{ big}}\approx
    \left(
    \frac{\Vert\be_0\Vert_1}
    {\Vert\be_0 - \be_1\Vert_1}
    \right)^{P+1}
    \,,
\end{equation}
taking the prior on $\eta$ to be uniform.
As another example, if we have homogeneous designs (viz. $\bS_1=\ldots=\bS_K$), then the operator $\T$ becomes asymptotically in $N$ a simple unweighted averaging: $\bar\T = \frac1{\bet^\top\mathbf{1}}[\eta_1\I,\ldots,\eta_K\I]$, such that the marginal asymptotic $\bet$ probabilities are approximately inversely proportional to $\Vert \be_0 - \frac1{\bet^\top\mathbf{1}}\sum_{\{k:\eta_k=1\}} \be_k\Vert_1^{P+1}$.

\section{\textit{Maximum A Posteriori} Analysis under Known $\bet$}
\label{sec:map}

This section studies the MAP estimator associated with our model.
We will contrast it with the two-stage approach of the Trans-Lasso.
To focus on this aspect, we will assume, only for this section, that hyperparameters are known. 
Assuming that $\bet$ is known means irrelevant datasets can simply be discarded and we may thus assume that $\bet = \mathbf{1}$ without loss of generality. 
Since $\sigma^2$ is known, we may scale the data such that it is unity.
Then, for a general source prior $P(\be_S)$, we obtain the following negative log density:
\begin{align*}
    \underset{\be_0,\ldots,\be_K}{\min} 
    \frac12\sum_{k=0}^K \Vert \y_k - \X_k\be_k\Vert_2^2 + 
    \lambda_t \Vert \T\be_S - \be_0\Vert_1 - \log P(\be_S) \,.
\end{align*}

\subsection{Trans-Lasso as an Approximate MAP-TRP}

We begin with a review of the \texttt{Trans-Lasso} algorithm, which is a two-step method. 
In the first step, a Lasso regression is fit to the concatenation of all source datasets to obtain an initial coefficient estimate $\tilde{\be}$.
Subsequently, the predictions from this coefficient on the target data, $\X_0\tilde{\be}$, are used in a second Lasso which regresses against the residuals from that model to form a correction term $\bd$.
Finally, the target coefficient estimate is given by the sum of those two coefficients.
Mathematically, it may be expressed as:
\begin{enumerate}
    \item 
    $ \tilde{\be} \gets \underset{\be}{\textrm{argmin}} \,\,
    \frac1{2}\sum_{k=1}^K \sum_{n=1}^{N_k} (y_{k,n} - \x_{k,n}^\top\be)^2 + \lambda_p \Vert \be \Vert_1
    \,,$ 
    \item
    $ 
    \hat{\bd} \gets \underset{\bd}{\textrm{argmin}}\,\,
    \frac1{2}\sum_{n=1}^{N_0} \left((y_{0,n} - \x_{0,n}^\top\tilde{\be}) - \x_{0,n}^\top\bd\right)^2 + \lambda_t \Vert \bd \Vert_1
    \,,$ 
    \item $\hat\be_0^{\textrm{TL}} \gets \tilde\be + \hat{\bd}$.
\end{enumerate}

To compare this against our MAP, we will use the improper prior $-\log P(\be_S) = C + \lambda_p\Vert\T \be_S\Vert_1$.
Note that this places the source penalty on $\T\be_S$ only rather than the entire $\be_S$ vector, and so differs from the prior used everywhere else in this article; we adopt it only in this subsection, because it is what establishes correspondence with the \texttt{Trans-Lasso}'s first step.
It leads to a cost decomposable as follows:
\begin{align*}
    & 
    \underset{\be_0,\ldots,\be_K}{\min} \,\,
    \frac12\sum_{k=0}^K \Vert \y_k - \X_k\be_k\Vert_2^2 + \lambda_t \Vert \T\be_S - \be_0\Vert_1 + \lambda_p\Vert\T\be_S\Vert_1 \,.
    \\ & = 
    \underset{\be_0,\bt}{\min} \,\,
    \left[
    \frac12\Vert \y_0 - \X_0\be_0\Vert_2^2 
    \right]
    + \lambda_t \Vert \bt - \be_0\Vert_1 
    +
    \left[
    \frac12\sum_{k=1}^K \Vert \y_k - \X_k\bt\Vert_2^2 
    + \lambda_p \Vert\bt\Vert_1 
    \right]
    \,,
\end{align*}
where $\bt := \T\be_S$ and to match the Trans-Lasso we have taken $\tau\to0$; see Appendix \ref{ap:map_cov} for details.
At first glance, this looks familiar to the fused lasso \citep{tibshirani2005sparsity}, however it is actually simpler since each parameter appears only in a single nonsmooth term, whereas in the fused lasso most parameters appear in two.

We will now undertake a sweep of block coordinate descent, initializing $\hat\be_0^{(0)}\gets \hat\bt^{(0)} \gets \bz$.
We start by minimizing with respect to $\bt$ while $\be_0=\mathbf{0}$.
Collecting relevant terms gives:
\begin{align*}
    & 
    \hat\bt^{(1)} \gets \argmin{\bt} \,\,
    \lambda_t \Vert \bt - \hat\be_0^{(0)}\Vert_1 
    +
    \frac12\sum_{k=1}^K \Vert \y_k - \X_k\bt\Vert_2^2 
    + \lambda_p \Vert\bt\Vert_1 
    \\ & =
    \argmin{\bt} \,\,
    \frac12\sum_{k=1}^K \Vert \y_k - \X_k\bt\Vert_2^2 
    + (\lambda_p+\lambda_t) \Vert\bt\Vert_1  \,.
\end{align*}
This is recognized as Step 1 of the Trans-Lasso, but with source coefficient given by $\lambda_p+\lambda_t$.
Subsequently, with $\bt$ fixed at $\hat\bt^{(1)}$, we optimize with respect to $\be_0$:
\begin{align*}
    \hat\be_0^{(1)}
    \gets
    \argmin{\be_0}
    \,\,
    \frac12\Vert \y_0 - \X_0\be_0\Vert_2^2 
    + \lambda_t \Vert \be_0 - \hat\bt^{(1)}\Vert_1 
    \,,
\end{align*}
which is recognized as identical to the second step of the transfer Lasso.

This reveals the sense in which a Trans-Lasso($\lambda_p$,$\lambda_t$) is an approximate solution of the TRP-MAP($\lambda_p-\lambda_t$,$\lambda_t$).
Intriguingly this only explains Trans-Lasso when source penalty exceeds transfer penalty.

This suggests a perspective of \texttt{Trans-Lasso} as a large source dataset approximation of conducting two steps of block-coordinate descent on the posterior density, as we discuss momentarily.
However, the question remains of whether there is some other Bayesian model whose MAP exactly matches the Trans-Lasso. 
We conjecture that the answer is negative.
This is because of the \texttt{Trans-Lasso}'s asymmetry: in our notation, it posits that $\be_0$ is close to $\bt$ when estimating $\be_0$, but does not use this information when estimating $\bt$. 
The Bayesian cannot abide this: if provided the prior information that $\be_0$ and $\bt$ are close, a rational actor must shrink estimates of each of them towards the other. 

Nevertheless, these procedures agree in the limit if \textit{either} the source data size \textit{or} the target data size diverges.
We formalize this in the following theorem, which omits the source penalty and holds uniformly over $\y_A$.
\begin{theorem}
    Assume the target and source design matrices are of full column rank, so that $\G{0}\succ\bz$ and $\sum_{k=1}^K\G{k}\succ\bz$, and let $\tau=\lambda_p=0$.
    Then, for any $\y_0,\ldots,\y_K$,
    \begin{align*}
    \left\Vert \hat\be_0^{\textrm{TRP}} - \hat\be_0^{\textrm{TL}}\right\Vert_2
    \leq 
    \frac{2\sqrt{P}\lambda_t}
    {
    \sqrt{\mineig(\G{0})} 
    \sqrt{\mineig\left(\sum_{k=1}^K \G{k} \right)}
    }
    \end{align*}

    \label{thm:tl_asymp}
    
\end{theorem}
\begin{proof}
    Appendix \ref{ap:tl_asymp}.
\end{proof}

\subsection{Computing the MAP Estimator for General $\X_A$}

The component  $\frac{\lambda_t}{\sigma}\Vert\be_0-\T\be_S\Vert_1$ of the prior density, being nonsmooth, rules out not only algorithms typically used for differentiable problems, such as gradient descent, Newton, or Quasi-Newton methods, but as we shall see, it also requires care when using the coordinate-descent methods typically used for nonsmoothly-penalized regression problems such as that implemented in the popular \texttt{glmnet} \citep{friedman2010regularization} or \texttt{ncvreg} \citep{ncvreg} R packages.
This section will discuss optimization strategies which can accommodate this nonsmoothness. 

\begin{figure}
    \centering
    \includegraphics[width=0.7\linewidth]{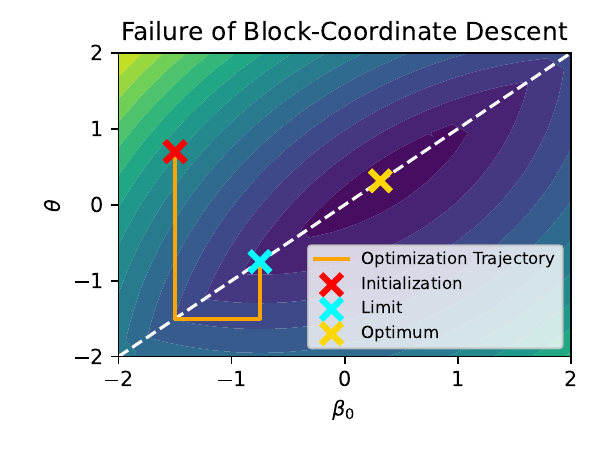}
    \caption{
    Failure of coordinate descent to maximize the posterior density in a simple synthetic $P=1$ problem.
    Here, $\hat\beta_0=0.75$, $\hat \theta = -0.3$, $s_0=0.3$, $s_z=0.5$, and $\lambda_t=0.5$, and the search is initialized at $\beta_0=-1.5$ and $\theta=0.7$.
    }
    \label{fig:bcd_fail}
\end{figure}

Off the bat, if we view \citet{li2022transfer} as approximately conducting the first two steps of a block-coordinate ascent, it is tempting to continue this procedure, alternating between $\be_0$ and $\bt$ steps up the posterior density.
Intriguingly, this does not converge to the global optimum despite the negative posterior log-density being a convex function (see Figure \ref{fig:bcd_fail}). 
This is because the nonsmoothness does not separate along each variable, as is the case in Lasso regression, which is necessary to guarantee convergence of (block) coordinate descent \citep{tseng2001convergence}. 
In fact, the cost surface induced by our log posterior density is similar to a classic counter-example used to show that coordinate descent does not reliably optimize convex functions without smoothness guarantees \citep{auslender1976optimisation}\footnote{Readers unfamiliar with these ideas will benefit from these excellent lecture notes: \url{https://www.cs.cmu.edu/~ggordon/10725-F12/slides/25-coord-desc.pdf}.}.
The same weakness precludes doing coordinate-descent on individual parameters, which is the prevailing technique for nonsmooth regression in statistics.

However, a simple transformation can put this problem into the form usually expected for Lasso regression. 
When the source datasets' relevances are known and only inference on $\be_0$ is desired, a change of variables demonstrates (Appendix \ref{ap:map_cov}) that, if $\tau=0$, we may concatenate all sources into a single source dataset $\X_1$, which we do for readability in the rest of this section.
Starting with our cost:
\begin{align}
    \min_{\be_0,\be_1}\,
    \frac12\Vert\y_0 - \X_0 \be_0\Vert_2^2 
    + \frac12\Vert\y_1 - \X_1\be_1\Vert_2^2
    + \lambda_t \Vert\be_0 - \be_1\Vert_1
    + \lambda_p \Vert \be_1 \Vert_1 \,,
\end{align}
we define $\z = \be_0 - \be_1$ and plug in for $\be_0 = \z + \be_1$, obtaining:
\begin{align}
    \min_{\z,\be_1}\,
    \frac12\Vert\y_0 - \X_0\be_1 - \X_0 \z\Vert_2^2 
    + \frac12\Vert\y_1 - \X_1\be_1\Vert_2^2
    + \lambda_t \Vert\z\Vert_1
    + \lambda_p \Vert \be_1 \Vert_1\,,
\end{align}
which may be rewritten as:
\begin{align}
&
    \min_{\z,\be_1} \,
    \frac12
    \left\Vert
    \begin{bmatrix}
        \y_0 \\ \y_1
    \end{bmatrix}
    -
    \begin{bmatrix}
        \X_0 & \X_0 \\
        \mathbf{0} & \X_1
    \end{bmatrix}
    \begin{bmatrix}
        \z \\ \be_1
    \end{bmatrix}
    \right\Vert_2^2
    + \lambda_t\Vert\z\Vert_1 + \lambda_p\Vert\be_1\Vert_1 
    \\ \iff & 
    \min_{\z,\tilde{\be_1}} \,
    \frac12
    \left\Vert
    \begin{bmatrix}
        \y_0 \\ \y_1
    \end{bmatrix}
    -
    \begin{bmatrix}
        \X_0 & \frac{\lambda_t}{\lambda_p}\X_0 \\
        \mathbf{0} & \frac{\lambda_t}{\lambda_p}\X_1
    \end{bmatrix}
    \begin{bmatrix}
        \z \\ \tilde{\be_1}
    \end{bmatrix}
    \right\Vert_2^2
    + \lambda_t \left\Vert\begin{bmatrix}
    \z \\ \tilde{\be_1}
    \end{bmatrix}\right\Vert_1\,.
\end{align}
This exhibits the problem as a standard Lasso regression.
Once fit, we calculate $\be_0^* = \z^* + \frac{\lambda_t}{\lambda_p}\tilde{\be}_1^*$ to obtain the MAP estimate for the target coefficients.

\subsection{Normal Means Analysis}

\label{sec:univariate}

The Normal Means setting is a commonly used meta-model of linear regression, and corresponds to either a univariate regression with no intercept or a multiple regression with orthogonal covariates.
In this section, we will leverage this simplified setting to gain greater insight into the MAP-TRP itself as well as its differences relative to the Trans-Lasso.
We therefore assume in this subsection that we observe two independent samples, $\y_1 \overset{iid}{\sim} N(\mu_1, \sigma_1^2)$ and $\y_0\overset{iid}{\sim} N(\mu_0, \sigma_0^2)$ with $\sigma_0^2,\sigma_1^2$ known.

This leads to the following problem:
\begin{equation}
    \underset{\mu_0,\mu_1\in\mathbb{R}}{\min}
    \frac{(\mu_0 - \bar{y}_0)^2}{2s_0} 
    + 
    \frac{(\mu_1-\bar{y}_1)^2}{2s_1}
    +
    \lambda_t |\mu_0-\mu_1| 
    + 
    \lambda_p |\mu_1|
    \label{eq:prox_cost}
    \,,
\end{equation}
where $s_0,s_1$ are the sampling variances of the sample means, i.e. $s_0 = \frac{\sigma_0^2}{N_0}$.

\subsubsection{The Case with $\lambda_p = 0$ (No Source Penalty)}

If $\lambda_p = 0$, a change of variables is sufficient to transform this into a Lasso-type problem (see Appendix \ref{ap:lamp0}), backtransformation from which gives the following optimum:
\begin{align}
    & \hat\mu_0^{\textrm{TRP}}(\y_0,\y_1;\lambda_t) = \frac{\frac{\bar{y}_0}{s_0} + \frac{\bar{y}_1}{s_1}}{\frac1{s_0} + \frac1{s_1}}
    +
    \frac{s_0}{s_0+s_1}
    \sgn(\bar{y}_0 - \bar{y}_1)
    \left(
    |\bar{y}_0 - \bar{y}_1|
    -
    (s_0+s_1)\lambda_t
    \right)^{+}
    \\ & :=
    \bar{w} + s_0 S
    \left(\frac{\bar{y}_0-\bar{y}_1}{s_0+s_1}, \lambda_t\right)    
    \,,
\end{align}
where $a^+ := \max(0,a)$ denotes the positive part of $a$, $\bar{w}$ gives the precision-weighted sample means, and $S:\mathbb{R}\times\mathbb{R}_+\to\mathbb{R}$ is the \textit{Soft Thresholding Operator} \citep{donoho1995noising}.
The MAP-TRP estimate in this context therefore starts at the precision-weighted mean and takes a step from that mean towards $\bar{y}_0$, the length of which is governed by $s_0$, $s_1$ and $\lambda_t$.

On the other hand, the \texttt{Trans-Lasso} without source penalty in this setting is given by:
\begin{align}
    \hat\mu_0^{\textrm{TL}}(\y_0,\y_1;\lambda_t) = \bar{y}_1 + \sgn(\bar{y}_0 - \bar{y}_1) 
    \left(
    |\bar{y}_0 - \bar{y}_1| - s_0 \lambda_t
    \right)^+ 
    =
    \bar{y}_1 + S(\bar{y}_0-\bar{y}_1, s_0\lambda_t)
    \,.
\end{align}

\begin{figure}
    \centering
    \includegraphics[width=0.96\linewidth]{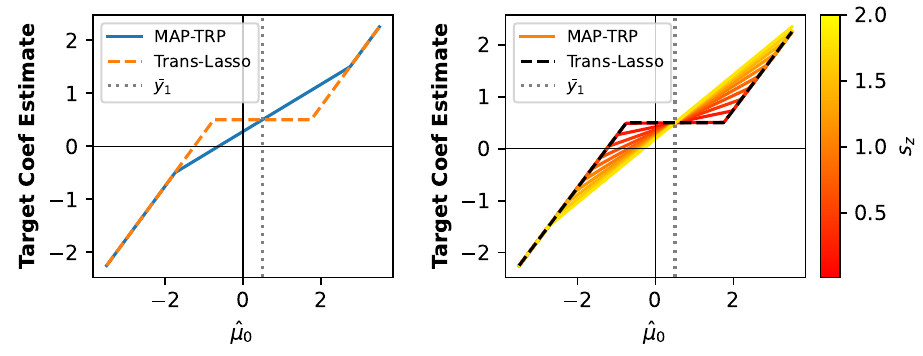}
    \caption{
    MAP-TRP vs TL for $\bar{y}_1 = \frac1{2}$ (gray line), $s_0=1.25$, $\lambda_t=1$.
    \textit{Left:}
    The MAP-TRP (solid blue line) disagrees with the modified \texttt{Trans-Lasso} method (dashed orange line) in the interval [-1.75, 2.75] when $s_1 = 1.25$.
    \textit{Right:}
    The MAP-TRP (solid autumn colors) for various $s_1\in[0,2]$ (red indicates small values, yellow large ones), which converges to the \texttt{Trans-Lasso} method indicated by the dashed black line for small $s_1$.
    }
    \label{fig:vsli}
\end{figure}

This makes it clear that the \texttt{Trans-Lasso} shrinks the target estimate towards the source estimate, while the MAP-TRP method shrinks the target estimate towards the precision-weighted average of both.
This is illustrated in Figure \ref{fig:vsli}, Left, which also illustrates that the two methods agree if $|\bar{y}_0-\bar{y}_1| > (s_0+s_1)\lambda_t$.
This fact itself implies that as the source variance vanishes, the two methods agree everywhere, i.e. $\underset{s_1\to 0}{\lim}  \hat\mu_0^{\textit{TRP}} = \hat\mu_0^{TL}$.
This is visualized in Figure \ref{fig:vsli}, Right.

We might intuitively think that, in finite samples, being able to use information from both datasets in producing the initial estimate would be better than just using the source sample for this purpose. 
The following result vindicates this intuition.
\begin{theorem}
    Fix any $\mu_0,\mu_1\in\mathbb{R}$ and $\sigma^2_0,\sigma^2_1,\lambda_t>0$, and let $\lambda_p = 0$.
    Let $\y_0\overset{iid}{\sim} N(\mu_0, \sigma_0^2)$ and $\y_1 \overset{iid}{\sim} N(\mu_1, \sigma_1^2)$.
    Then:
    \begin{align*}
        \underset{\y_0,\y_1}{\mathbb{E}}
        \left[
        \left(
        \hat{\mu}^{\textrm{TRP}}_0(\y_0,\y_1; \lambda_t)
        -\mu_0
        \right)^2
        \right]
        <
        \underset{\y_0,\y_1}{\mathbb{E}}
        \left[
        \left(
        \hat{\mu}^{\textrm{TL}}_0(\y_0,\y_1;\lambda_t)
        -\mu_0
        \right)^2
        \right] \,.
    \end{align*}
\end{theorem}
\begin{proof}
    First note the risks are equal when $\lambda_t=0$, then demonstrate that the derivative of their difference with respect to $\lambda_t$ is everywhere positive.
    See Appendix \ref{ap:risk}.
\end{proof}
This result demonstrates that the TRP dominates the TL in estimating $\mu_0$ under $\ell_2$ loss when $\lambda_p=0$ in the normal means setting.

\textit{Remark:} Since this is true for any setting of $\lambda_t$, it is also true for the optimal setting of each; let $R_{TL}(\mu_0,\mu_1,s_0,s_1,\lambda_t)$ give the MSE of the TL estimator under the given generative parameters and that regularization parameter, and similarly for $R_{TRP}(\mu_0,\mu_1,s_0,s_1,\lambda_t)$. 
Then,
$
\forall \mu_0,\mu_1\in\mathbb{R},s_0,s_1>0:
$
\begin{align*}
    \min_{\lambda_t} R_{TL}(\mu_0,\mu_1,s_0,s_1,\lambda_t) > 
    R_{TRP}(\mu_0,\mu_1,s_0,s_1,\lambda_t^*)
    \geq
    \min_{\lambda_t} R_{TRP}(\mu_0,\mu_1,s_0,s_1,\lambda_t)  \,,
\end{align*}
where we have denoted the optimal $\lambda_t$ value for the Trans Lasso as $\lambda_t^*$.

\paragraph{Non-extension to the Multivariate Case.}
Though simulation studies reveal that for many multivariate settings the Trans-Lasso has greater estimation MSE than the MAP-TRP, this is not a guarantee as in the normal means case.
One such example of non-dominance occurs in the large $\lambda_t$ setting, for which both methods will proffer fused coefficients.
For Trans-Lasso, this involves using $\hat\be_1^{\textrm{OLS}}$ as an estimate of $\be_0$ while MAP-TRP uses 
$
\left(
\G{0}+\G{1}
\right)^{-1}
\left(
\G{0}\hat\be^{\textrm{OLS}}_0
+
\G{1}\hat\be^{\textrm{OLS}}_1
\right)
$
.
The bias for these methods is given as
$\be_1 - \be_0$ 
and 
$
\left(
\G{0}+\G{1}
\right)^{-1}
\G{1}(\be_1 - \be_0)
$, respectively.
If the prefactor for the TRP bias has spectral norm greater than 1, this means there is a contrast $\be_1-\be_0$ for which the TRP bias exceeds the TL bias, and choosing a sufficiently small variance therefore gives advantage to the Trans-Lasso.
For example, the $2\times2$ matrices 
$\G{0} = \mathbf{1}\mathbf{1}^\top+\epsilon\I$ and $\G{1} = \diag([1, \epsilon])$ exhibit this effect for sufficiently small $\epsilon$, such as $\epsilon=\frac1{10}$.

\subsubsection{Normal Means under the General $\lambda_p$ Case}

Next, we will study the general solution to Problem \ref{eq:prox_cost}, staying within the Normal Means setting.
As demonstrated in Appendix \ref{ap:lamp_big_prox}, it has solution given by:
\begin{align*}
    &
    z \gets S\left(\frac{s_{1} \bar{y}_0 + s_{0} \bar{y}_1}{s_0+s_{1}}, \frac{\lambda_p}{\frac{1}{s_0}+\frac{1}{s_1}}\right) \,;
    \\ &
    \hat\mu^{TRP}_0(\y_0,\y_1;\lambda_t,\lambda_p) \gets z + S\left(\bar{y}_0 - z, s_0\lambda_t\right)
    \,,
\end{align*}
that is, the MAP-TRP starts at a shrunken precision-weighted mean of the two samples, and subsequently moves in the direction of the target mean.
On the other hand, in this context the Trans-Lasso gives:
\begin{align*}
    & z \gets 
    S(\bar{y}_1, \lambda_p s_1)\,; 
    \\ & 
    \hat\mu_0^{TL}(\y_0,\y_1;\lambda_t,\lambda_p) \gets z + S(\bar{y}_0 - z, s_0\lambda_t)\,,
\end{align*}
which instead starts at a shrunken source mean, though it intriguingly shares its modification step with the MAP-TRP.

Unlike the $\lambda_p=0$ case, we do not observe that the \texttt{Trans-Lasso} is dominated by the MAP-TRP for all settings of $\lambda_t, \lambda_p$.
However, we do find, numerically, that holding $\lambda_t$ fixed and optimizing the risk over $\lambda_p$ leads to empirical dominance across 2,000 randomly sampled settings of $\mu_0,\mu_1 \sim N(0,1)$ and $s_0,s_1,\lambda_t\sim \Gamma(1,1)$.
This suggests that the non-dominance may not be intrinsic to each method, but rather due to a parameterization disagreement. 
It leads us to the following conjecture.
\begin{conjecture}
    Fix any $\mu_0,\mu_1\in\mathbb{R}$ and $\sigma^2_0,\sigma^2_1,\lambda_t>0$.
    Let $\y_0\overset{iid}{\sim} N(\mu_0, \sigma_0^2)$ and $\y_1 \overset{iid}{\sim} N(\mu_1, \sigma_1^2)$.
    Then:
    \begin{align*}
        \min_{\lambda_p}
        \underset{\y_0,\y_1}{\mathbb{E}}
        \left[
        \left(
        \hat{\mu}^{TRP}_0(\y_0,\y_1; \lambda_t,\lambda_p)
        -\mu_0
        \right)^2
        \right]
        <
        \min_{\lambda_p}
        \underset{\y_0,\y_1}{\mathbb{E}}
        \left[
        \left(
        \hat{\mu}^{TL}_0(\y_0,\y_1; \lambda_t,\lambda_p)
        -\mu_0
        \right)^2
        \right] \,.
    \end{align*}
\end{conjecture}

\section{Application}
\label{sec:app}

Generally following \citet{li2022transfer}, we study the performance of our method in the context of Genotype-Tissue Expression (GTEx) data.
In particular, we use the Adult GTEx data giving Gene Expression Transcripts Per Million (TPM) obtained from the GTEx Portal\footnote{\url{https://www.gtexportal.org/home/downloads/adult-gtex/bulk_tissue_expression}}, and focus on the genes from the central nervous system contained in Module 137.
The response variable is the JAM3 expression, and we use the other $P=399$ gene expressions as predictor variables.
There are 37 tissues, the number of observations for which varied from 107 to 2,697.

We used this dataset to establish a transfer learning benchmark in which each tissue constitutes a dataset.
At each iteration, we randomly choose one tissue to serve as the target dataset.
Then, of the remaining 36, we randomly choose some subset of size $K$ of them to serve as source datasets, which allows us to tune the extent to which our setting is constrained by source data by tuning $K$.
We randomly hold out $20\%$ of the target data to serve as test data, and we evaluate the performance of various methods in predicting these held-out data in terms of MSE.
We repeated this procedure 100 times for $K\in\{4,8,16,32\}$.

We implement our proposed TRP method using the Laplace prior with $\gamma = \frac{\tau}{2}\Vert\cdot\Vert_2^2$.
We use $10{,}000$ MCMC iterations with a burnin of $2{,}000$, and use $L=5$ temperatures for parallel tempering.
We compare this against two frequentist implementations of transfer learning, namely an implementation of \texttt{Trans-Lasso} derived from the codebase accompanying the article presenting that method\footnote{\url{https://github.com/saili0103/TransLasso/tree/main}} as well as to the \texttt{glmtrans} package \citep{tian2023transfer} using the Gaussian error family.
In addition, we also compare to the BLAST method of \citet{jamshidian2025bayesian} using the repository accompanying their article.
For context, we also include an OLS using  all source datasets and the target dataset combined (\texttt{Pooled OLS}) as well as a Lasso regression fit only to the target dataset (\texttt{Target Lasso}), estimated using \texttt{glmnet} using the standard cross-validation procedure to choose the penalty.

\begin{figure}
    \centering
    \includegraphics[width=0.75\linewidth]{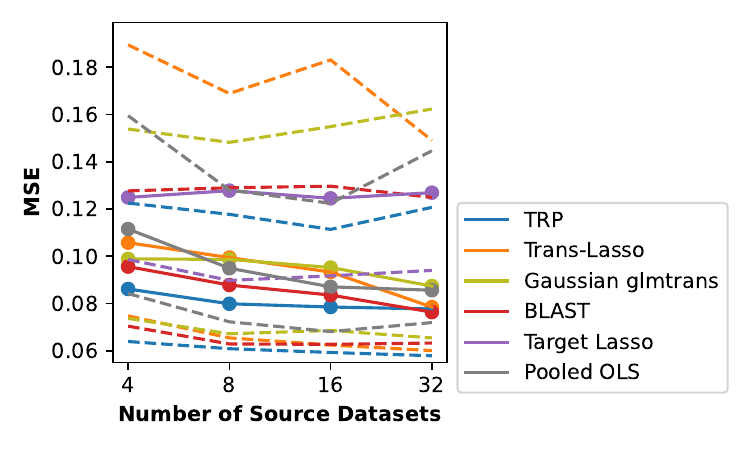}
    \caption{Comparison of predictive performance. Solid line gives median MSE and dotted lines give 25th and 75th percentile of MSE distribution over 100 repetitions. The 75th percentile for the \texttt{Target Lasso} method is off chart high at about 0.25.}
    \label{fig:gene}
\end{figure}

The results are given in Figure \ref{fig:gene}, which shows the out-of-sample MSE distribution aggregated across all tissues.
We see that the TRP outperforms the other methods, especially for small $K$, in terms of median performance.
Intriguingly, the 75th percentile remains lower for the TRP relative to competing methods in the $K=32$ setting, suggesting milder predictive behavior in unfavorable situations.
When using a small number of auxiliary datasets, we find that while \texttt{Trans-Lasso} has only somewhat better median-case behavior than a simple pooled OLS regression, it actually has worse upper quartile behavior.

\section{Discussion}
\label{sec:conc}

We end this article by providing a summary of our contributions, a discussion of the method's limitations, and the prospects for future work.

\subsection{Summary}
We began by showing how frequentist transfer learning methods have behavior qualitatively different than Bayesian hierarchical models, which can lead to surprising differences as illustrated by our toy dataset based on Simpson's paradox. 
We then proposed the Total Risk Prior, endowing a formal Bayesian model with the behavior of frequentist transfer learning, and which can maximally leverage small source datasets. 
Next, we discussed a source-selection consistency property, before  demonstrating that the method improves on the \texttt{Trans-Lasso} algorithm of \citet{li2022transfer} both in terms of normal means MSE and predictive performance on a genetics problem.

\subsection{Limitations}

The primary limitation of our methodology is its computationally intensive nature. 
We were only able to apply it to our moderate setting of $36$ source datasets with at most a few thousand observations each and about 400 predictor variables after significant attention to computational efficiency, and it still took about 5 hours to run the $10,000$ iteration chain on a 1080ti GPU.
While trifling relative to the computational expense of performing inference with Large Language Models, for example, it is still expensive for a linear model.
This makes it most suitable for high-stakes applications with severe data limitations where it is worth the flops to extract every bit of information.

\subsection{Future Work}

A clear avenue for future work is the extension of the idea of using a risk minimizer as a prior hyperparameter to settings beyond the linear models.
While the general principle is already clear, there remains significant work to be done to achieve a computationally tractable algorithm and to elucidate the mathematical properties of this procedure.
We suspect that there is structure to be uncovered in, for instance, the application of a TRP in the context of a Generalized Linear Model.
Additionally, while we presently stand some way short of applying this prior to a Bayesian neural network, the potential to go full circle by applying this methodology to deep learning is also quite tempting.
There remain interesting mathematical questions about the behavior of the TRP with linear models as presented in this article.
Even in the normal means model, Conjecture 1 remains open.

\subsubsection*{Data Availability Statement}
The processed data is available in the Supplementary Material as a zip file.

\subsubsection*{Generative AI Statement}
In this work, we used generative AI tools to assist in developing the mathematical claims and for checking mistakes in the writeup.
The mathematical proofs presented in this work were all written manually and are based in part on output from various commercial LLMs including Claude Fable 5.1 and ChatGPT 6.1 Sol.
In all cases, the exact arguments presented here were synthesized by the authors after iteration with multiple LLM generated proofs and independent reflection.
We did not use AI for writing the manuscript narrative, for developing the core concept, or for writing the original code, but did use AI bug checking and documentation.
We take full responsibility for the final content of this work.

\bibliography{main}

@article{suder2023bayesian,
  title={Bayesian transfer learning},
  author={Suder, Piotr M and Xu, Jason and Dunson, David B},
  journal={arXiv preprint arXiv:2312.13484},
  year={2023}
}

@article{abba2024bayesian,
  title={A {B}ayesian shrinkage estimator for transfer learning},
  author={Abba, Mohamed A and Williams, Jonathan P and Reich, Brian J},
  journal={arXiv preprint arXiv:2403.17321},
  year={2024}
}

@article{hickey2024transfer,
  title={Transfer learning with uncertainty quantification: Random effect calibration of source to target ({REC}a{ST})},
  author={Hickey, Jimmy and Williams, Jonathan P and Hector, Emily C},
  journal={Journal of Machine Learning Research},
  volume={25},
  number={338},
  pages={1--40},
  year={2024}
}

@inproceedings{raina2006constructing,
  title={Constructing informative priors using transfer learning},
  author={Raina, Rajat and Ng, Andrew Y and Koller, Daphne},
  booktitle={Proceedings of the 23rd international conference on Machine learning},
  pages={713--720},
  year={2006}
}

@article{zhang2024covariate,
  title={Covariate-Elaborated Robust Partial Information Transfer with Conditional Spike-and-Slab Prior},
  author={Zhang, Ruqian and Zhang, Yijiao and Qu, Annie and Zhu, Zhongyi and Shen, Juan},
  journal={arXiv preprint arXiv:2404.03764},
  year={2024}
}

@article{li2022transfer,
  title={Transfer learning for high-dimensional linear regression: Prediction, estimation and minimax optimality},
  author={Li, Sai and Cai, T Tony and Li, Hongzhe},
  journal={Journal of the Royal Statistical Society Series B: Statistical Methodology},
  volume={84},
  number={1},
  pages={149--173},
  year={2022},
  publisher={Oxford University Press}
}

@article{park2008bayesian,
  title={The {B}ayesian lasso},
  author={Park, Trevor and Casella, George},
  journal={Journal of the american statistical association},
  volume={103},
  number={482},
  pages={681--686},
  year={2008},
  publisher={Taylor \& Francis}
}

@article{pramanik2025modeling,
  title={Modeling structure and country-specific heterogeneity in misclassification matrices of verbal autopsy-based cause of death classifiers},
  author={Pramanik, Sandipan and Zeger, Scott and Blau, Dianna and Datta, Abhirup},
  journal={The Annals of Applied Statistics},
  volume={19},
  number={2},
  pages={1214--1239},
  year={2025},
  publisher={Institute of Mathematical Statistics}
}

@article{hans2009bayesian,
  title={Bayesian lasso regression},
  author={Hans, Chris},
  journal={Biometrika},
  volume={96},
  number={4},
  pages={835--845},
  year={2009},
  publisher={Oxford University Press}
}

@article{swendsen1986replica,
  title={Replica Monte Carlo simulation of spin-glasses},
  author={Swendsen, Robert H and Wang, Jian-Sheng},
  journal={Physical review letters},
  volume={57},
  number={21},
  pages={2607},
  year={1986},
  publisher={APS}
}

@article{geyer1991computing,
  title={Computing science and statistics: Proceedings of the 23rd Symposium on the Interface},
  author={Geyer, Charles J and others},
  journal={American Statistical Association, New York},
  volume={156},
  year={1991}
}

@article{robert2021rao,
  title={Rao-Blackwellization in the MCMC era},
  author={Robert, Christian P and Roberts, Gareth O},
  journal={arXiv preprint arXiv:2101.01011},
  year={2021}
}

@article{carvalho2010horseshoe,
  title={The horseshoe estimator for sparse signals},
  author={Carvalho, Carlos M and Polson, Nicholas G and Scott, James G},
  journal={Biometrika},
  pages={465--480},
  year={2010},
  publisher={JSTOR}
}

@article{robbins1951stochastic,
  title={A stochastic approximation method},
  author={Robbins, Herbert and Monro, Sutton},
  journal={The annals of mathematical statistics},
  pages={400--407},
  year={1951},
  publisher={JSTOR}
}

@article{makalic2015simple,
  title={A simple sampler for the horseshoe estimator},
  author={Makalic, Enes and Schmidt, Daniel F},
  journal={IEEE Signal Processing Letters},
  volume={23},
  number={1},
  pages={179--182},
  year={2015},
  publisher={IEEE}
}

@article{li2023targeting,
  title={Targeting underrepresented populations in precision medicine: A federated transfer learning approach},
  author={Li, Sai and Cai, Tianxi and Duan, Rui},
  journal={The annals of applied statistics},
  volume={17},
  number={4},
  pages={2970},
  year={2023}
}

@article{shang2024self,
  title={Self-starting monitoring schemes for small-sample poisson profiles based on transfer learning},
  author={Shang, Yanfen and Lu, Chang and Li, Longhui and He, Shuguang},
  journal={Computers \& Industrial Engineering},
  volume={192},
  pages={110262},
  year={2024},
  publisher={Elsevier}
}

@article{hu2019statistical,
  title={A statistical framework for cross-tissue transcriptome-wide association analysis},
  author={Hu, Yiming and Li, Mo and Lu, Qiongshi and Weng, Haoyi and Wang, Jiawei and Zekavat, Seyedeh M and Yu, Zhaolong and Li, Boyang and Gu, Jianlei and Muchnik, Sydney and others},
  journal={Nature genetics},
  volume={51},
  number={3},
  pages={568--576},
  year={2019},
  publisher={Nature Publishing Group US New York}
}

@inproceedings{daume2007frustratingly,
  title={Frustratingly Easy Domain Adaptation},
  author={Daum{\'e} III, Hal},
  booktitle={Proceedings of the 45th Annual Meeting of the Association of Computational Linguistics},
  pages={256--263},
  year={2007}
}

@inproceedings{wang2018deep,
  title={Deep reinforcement learning with knowledge transfer for online rides order dispatching},
  author={Wang, Zhaodong and Qin, Zhiwei and Tang, Xiaocheng and Ye, Jieping and Zhu, Hongtu},
  booktitle={2018 IEEE International Conference on Data Mining (ICDM)},
  pages={617--626},
  year={2018},
  organization={IEEE}
}

@article{hajiramezanali2018bayesian,
  title={Bayesian multi-domain learning for cancer subtype discovery from next-generation sequencing count data},
  author={Hajiramezanali, Ehsan and Zamani Dadaneh, Siamak and Karbalayghareh, Alireza and Zhou, Mingyuan and Qian, Xiaoning},
  journal={Advances in Neural Information Processing Systems},
  volume={31},
  year={2018}
}

@article{zhang2022survey,
  title={A survey on negative transfer},
  author={Zhang, Wen and Deng, Lingfei and Zhang, Lei and Wu, Dongrui},
  journal={IEEE/CAA Journal of Automatica Sinica},
  volume={10},
  number={2},
  pages={305--329},
  year={2022},
  publisher={IEEE}
}

@article{zhuang2020comprehensive,
  title={A comprehensive survey on transfer learning},
  author={Zhuang, Fuzhen and Qi, Zhiyuan and Duan, Keyu and Xi, Dongbo and Zhu, Yongchun and Zhu, Hengshu and Xiong, Hui and He, Qing},
  journal={Proceedings of the IEEE},
  volume={109},
  number={1},
  pages={43--76},
  year={2020},
  publisher={Ieee}
}

@article{pan2009survey,
  title={A survey on transfer learning},
  author={Pan, Sinno Jialin and Yang, Qiang},
  journal={IEEE Transactions on knowledge and data engineering},
  volume={22},
  number={10},
  pages={1345--1359},
  year={2009},
  publisher={IEEE}
}

@article{weiss2016survey,
  title={A survey of transfer learning},
  author={Weiss, Karl and Khoshgoftaar, Taghi M and Wang, DingDing},
  journal={Journal of Big data},
  volume={3},
  number={1},
  pages={9},
  year={2016},
  publisher={Springer}
}

@article{tian2023transfer,
  title={Transfer learning under high-dimensional generalized linear models},
  author={Tian, Ye and Feng, Yang},
  journal={Journal of the American Statistical Association},
  volume={118},
  number={544},
  pages={2684--2697},
  year={2023},
  publisher={Taylor \& Francis}
}

@article{li2024estimation,
  title={Estimation and inference for high-dimensional generalized linear models with knowledge transfer},
  author={Li, Sai and Zhang, Linjun and Cai, T Tony and Li, Hongzhe},
  journal={Journal of the American Statistical Association},
  volume={119},
  number={546},
  pages={1274--1285},
  year={2024},
  publisher={Taylor \& Francis}
}

@article{li2023transfer,
  title={Transfer learning in large-scale gaussian graphical models with false discovery rate control},
  author={Li, Sai and Cai, T Tony and Li, Hongzhe},
  journal={Journal of the American Statistical Association},
  volume={118},
  number={543},
  pages={2171--2183},
  year={2023},
  publisher={Taylor \& Francis}
}

@misc{jax2018github,
  author = {James Bradbury and Roy Frostig and Peter Hawkins and Matthew James Johnson and Chris Leary and Dougal Maclaurin and George Necula and Adam Paszke and Jake Vander{P}las and Skye Wanderman-{M}ilne and Qiao Zhang},
  title = {{JAX}: composable transformations of {P}ython+{N}um{P}y programs},
  url = {http://github.com/jax-ml/jax},
  version = {0.3.13},
  year = {2018},
}

@article{george1993variable,
  title={Variable selection via Gibbs sampling},
  author={George, Edward I and McCulloch, Robert E},
  journal={Journal of the American Statistical Association},
  volume={88},
  number={423},
  pages={881--889},
  year={1993},
  publisher={Taylor \& Francis}
}

@article{green1995reversible,
  title={Reversible jump Markov chain Monte Carlo computation and {B}ayesian model determination},
  author={Green, Peter J},
  journal={Biometrika},
  volume={82},
  number={4},
  pages={711--732},
  year={1995},
  publisher={Oxford University Press}
}

@article{metropolis1953equation,
  title={Equation of state calculations by fast computing machines},
  author={Metropolis, Nicholas and Rosenbluth, Arianna W and Rosenbluth, Marshall N and Teller, Augusta H and Teller, Edward},
  journal={The journal of chemical physics},
  volume={21},
  number={6},
  pages={1087--1092},
  year={1953},
  publisher={American Institute of Physics}
}

@article{geman1984stochastic,
  title={Stochastic relaxation, Gibbs distributions, and the {B}ayesian restoration of images},
  author={Geman, Stuart and Geman, Donald},
  journal={IEEE Transactions on pattern analysis and machine intelligence},
  number={6},
  pages={721--741},
  year={1984},
  publisher={IEEE}
}

@article{gelfand1990sampling,
  title={Sampling-based approaches to calculating marginal densities},
  author={Gelfand, Alan E and Smith, Adrian FM},
  journal={Journal of the American statistical association},
  volume={85},
  number={410},
  pages={398--409},
  year={1990},
  publisher={Taylor \& Francis}
}

@article{tierney1994markov,
  title={Markov chains for exploring posterior distributions},
  author={Tierney, Luke},
  journal={the Annals of Statistics},
  pages={1701--1728},
  year={1994},
  publisher={JSTOR}
}

@book{brooks2011handbook,
  title={Handbook of markov chain monte carlo},
  author={Brooks, Steve and Gelman, Andrew and Jones, Galin and Meng, Xiao-Li},
  year={2011},
  publisher={CRC press}
}

@article{bhattacharya2016fast,
  title={Fast sampling with Gaussian scale mixture priors in high-dimensional regression},
  author={Bhattacharya, Anirban and Chakraborty, Antik and Mallick, Bani K},
  journal={Biometrika},
  pages={asw042},
  year={2016},
  publisher={Oxford University Press}
}

@article{friedman2010regularization,
  title={Regularization paths for generalized linear models via coordinate descent},
  author={Friedman, Jerome H and Hastie, Trevor and Tibshirani, Rob},
  journal={Journal of statistical software},
  volume={33},
  number={1},
  pages={1--22},
  year={2010}
}

@article{tibshirani2005sparsity,
  title={Sparsity and smoothness via the fused lasso},
  author={Tibshirani, Robert and Saunders, Michael and Rosset, Saharon and Zhu, Ji and Knight, Keith},
  journal={Journal of the Royal Statistical Society Series B: Statistical Methodology},
  volume={67},
  number={1},
  pages={91--108},
  year={2005},
  publisher={Oxford University Press}
}

@book{auslender1976optimisation,
 title={Optimisation: M\'ethodes Num\'eriques},
 author={Auslender, A.},
 year={1976},
 publisher={Masson}
}

@article{tseng2001convergence,
  title={Convergence of a block coordinate descent method for nondifferentiable minimization},
  author={Tseng, Paul},
  journal={Journal of optimization theory and applications},
  volume={109},
  number={3},
  pages={475--494},
  year={2001},
  publisher={Springer}
}

@Article{ncvreg,
author = {Patrick Breheny and Jian Huang},
title = {Coordinate descent algorithms for nonconvex penalized
  regression, with applications to biological feature selection},
journal = {Annals of Applied Statistics},
year = {2011},
volume = {5},
pages = {232--253},
number = {1},
doi = {10.1214/10-AOAS388},
url = {https://doi.org/10.1214/10-AOAS388},
}

@article{bull2023hierarchical,
  title={Hierarchical {B}ayesian modeling for knowledge transfer across engineering fleets via multitask learning},
  author={Bull, Lawrence A and Di Francesco, Domenic and Dhada, Maharshi and Steinert, Olof and Lindgren, Tony and Parlikad, Ajith Kumar and Duncan, Andrew B and Girolami, Mark},
  journal={Computer-Aided Civil and Infrastructure Engineering},
  volume={38},
  number={7},
  pages={821--848},
  year={2023},
  publisher={Wiley Online Library}
}

@article{zhang2021survey,
  title={A survey on multi-task learning},
  author={Zhang, Yu and Yang, Qiang},
  journal={IEEE transactions on knowledge and data engineering},
  volume={34},
  number={12},
  pages={5586--5609},
  year={2021},
  publisher={IEEE}
}

@article{carlin1991inference,
  title={Inference for nonconjugate Bayesian models using the Gibbs sampler},
  author={Carlin, Bradley P and Polson, Nicholas G},
  journal={Canadian Journal of statistics},
  volume={19},
  number={4},
  pages={399--405},
  year={1991},
  publisher={Wiley Online Library}
}

@article{andrews1974scale,
  title={Scale mixtures of normal distributions},
  author={Andrews, David F and Mallows, Colin L},
  journal={Journal of the Royal Statistical Society: Series B (Methodological)},
  volume={36},
  number={1},
  pages={99--102},
  year={1974},
  publisher={Wiley Online Library}
}

@book{rockafellar1970convex,
 ISBN = {9780691015866},
 URL = {http://www.jstor.org/stable/j.ctt14bs1ff},
 author = {R. Tyrrell Rockafellar},
 publisher = {Princeton University Press},
 title = {Convex Analysis},
 urldate = {2026-08-05},
 year = {1970}
}

@article{hjort2011asymptotics,
  title={Asymptotics for minimisers of convex processes},
  author={Hjort, Nils Lid and Pollard, David},
  journal={arXiv preprint arXiv:1107.3806},
  year={1993}
}

@article{jamshidian2025bayesian,
  title={Bayesian Transfer Learning for High-Dimensional Linear Regression via Adaptive Shrinkage},
  author={Jamshidian, Parsa and Telesca, Donatello},
  journal={arXiv preprint arXiv:2510.03449},
  year={2025}
}

@article{lai2026bayesian,
  title={Bayesian transfer learning for enhanced estimation and inference},
  author={Lai, Daoyuan and Madrid Padilla, Oscar Hernan and Li, Xiang and Gu, Tian},
  journal={Journal of the American Statistical Association},
  number={just-accepted},
  pages={1--28},
  year={2026},
  publisher={Taylor \& Francis}
}

@article{liang2008mixtures,
  title={Mixtures of g priors for Bayesian variable selection},
  author={Liang, Feng and Paulo, Rui and Molina, German and Clyde, Merlise A and Berger, Jim O},
  journal={Journal of the American Statistical Association},
  volume={103},
  number={481},
  pages={410--423},
  year={2008},
  publisher={Taylor \& Francis}
}

@article{zellner1986assessing,
  title={On assessing prior distributions and Bayesian regression analysis with g-prior distributions},
  author={Zellner, Arnold},
  journal={Bayesian inference and decision techniques},
  year={1986},
  publisher={Elsevier Science}
}

@article{presicce2026bayesian,
  title={Bayesian transfer learning for artificially intelligent geospatial systems: A predictive stacking approach},
  author={Presicce, Luca and Banerjee, Sudipto},
  journal={Journal of Machine Learning Research},
  volume={27},
  number={196},
  pages={1--60},
  year={2026}
}

@article{donoho1995noising,
  title={De-noising by soft-thresholding},
  author={Donoho, David L},
  journal={IEEE transactions on information theory},
  volume={41},
  number={3},
  pages={613--627},
  year={1995},
  publisher={IEEE}
}

\newpage

\appendix

\section{Outline of Appendices}

Throughout the appendices we suppress the dependence of the transfer matrix on
the inclusion indicators, writing $\T$ for $\T_{\bet}$, and we write
$\X_S := \diag(\X_1,\ldots,\X_K)$ for the block-diagonal design matrix of the
source datasets alone, so that $\G{S} = \diag(\G{1},\ldots,\G{K})$.
We write $\hbe_k := \Gi{k}\X_k^\top\y_k$ for the $k$th OLS estimate,
$\hat\bs := [\hbe_1^\top,\ldots,\hbe_K^\top]^\top$ for the stacked source OLS
estimates, and $\hat\bt := \T\hat\bs$.

We begin with Appendix \ref{ap:tmath} which derives some basic linear algebraic facts about the transfer matrix which will be useful in subsequent sections.
Next is Appendix \ref{ap:postasymp}, which discusses the asymptotic posterior of TRP for a Gaussian linear regression.
Appendix \ref{ap:map_cov} goes into detail about the change of variables used to reduce multiple sources to a single source when $\bet$ is known, while Appendix \ref{ap:tl_asymp} demonstrates asymptotic equivalence of the TRP-MAP and \texttt{Trans-Lasso}.
Appendices \ref{ap:prox} and \ref{ap:risk} give full details regarding the normal means comparison of the TRP-MAP with \texttt{Trans-Lasso} in the finite sample setting.

\section{Basic Computations with $\T$ (for $\tau=0$)}
\label{ap:tmath}

In several places below, we will benefit from acknowledging some elementary
linear algebraic properties of $\T$.
The projection onto $\T$'s rowspace may be shown to be:
\begin{align}
    & \mathbf{P}_{\T}
    = \T^\top(\T\T^\top)^{-1}\T
    =
    \begin{bmatrix}
        \G{1} \\
        \vdots\\
        \G{K}
    \end{bmatrix}
    \left(\sum_{k=1}^K (\G{k})^2 \right)^{-1}
    \begin{bmatrix}
        \G{1} &
        \ldots&
        \G{K}
    \end{bmatrix}
    \\ & =
    \begin{bmatrix}
        \G{1}
        \left(\sum_{k=1}^K (\G{k})^2 \right)^{-1}
        \G{1}
        &
        \ldots
        &
        \G{1}
        \left(\sum_{k=1}^K (\G{k})^2 \right)^{-1}
        \G{K}
        \\
        \vdots & \ddots & \vdots
        \\
        \G{K}
        \left(\sum_{k=1}^K (\G{k})^2 \right)^{-1}
        \G{1}
        &
        \ldots
        &
        \G{K}
        \left(\sum_{k=1}^K (\G{k})^2 \right)^{-1}
        \G{K}
    \end{bmatrix}
    \,.
\end{align}

Also, if we take the vector of OLS estimates for each dataset,
\begin{equation}
    \hat{\bs} :=
    \begin{bmatrix}
        \Gi{1}\X_1^\top \y_1 \\
        \vdots \\
        \Gi{K}\X_K^\top \y_K
    \end{bmatrix}\,,
\end{equation}
and hit it with the transfer matrix with $\tau=0$, the result is the overall
combined error minimizer:
\begin{equation}
    \T\hat\bs =
    \left(\sum_{k=1}^K \G{k} \right)^{-1}
    \begin{bmatrix}
        \G{1} & \ldots &
        \G{K}
    \end{bmatrix}
    \begin{bmatrix}
        \Gi{1}\X_1^\top \y_1 \\
        \vdots \\
        \Gi{K}\X_K^\top \y_K
    \end{bmatrix}
    =
    \left(\sum_{k=1}^K \G{k} \right)^{-1}
    \sum_{k=1}^K \X_k^\top\y_k \,.
\end{equation}

We also have that:
\begin{align}
    \T^\dagger =
    \begin{bmatrix}
        \G{1} \\
        \vdots \\
        \G{K}
    \end{bmatrix}
    \left(
    \sum_{k=1}^K (\G{k})^2
    \right)^{-1}
    \left(
    \sum_{k=1}^K \G{k}
    \right) \,,
\end{align}
as it is easily shown that this purported pseudo-inverse satisfies the standard
properties, namely that $\T\T^\dagger = \I$ and $\T^\dagger\T = \mathbf{P}_{\T}$
are both symmetric, and that consequently
$\T\T^\dagger\T = \mathbf{P}_{\T}\T = \T$ and
$\T^\dagger\T\T^\dagger = \T^\dagger\I = \T^\dagger$.

Furthermore, defining:
\begin{align}
    \W = \begin{bmatrix}
        \I \\ \vdots \\ \I
    \end{bmatrix} \in\mathbb{R}^{KP\times P} \,,
\end{align}
we notice that this is a right inverse of $\T$, though of course not the
Moore-Penrose pseudo-inverse as given above.

Given this notation, we can also express:
\begin{align}
    & \T = \left(
    \W^\top\G{S}\W
    \right)^{-1}
    \W^\top\G{S}
    \\
    & \T^\dagger =
    \G{S}\W
    \left(
    \W^\top(\G{S})^2\W
    \right)^{-1}
    \W^\top\G{S}\W\,.
\end{align}

\section{Posterior Details}\label{ap:postasymp}

We begin by reprinting the assumptions made in the body.
We use a bar to denote a limiting quantity.
In the case of identified parameters, such as $\ba$ under our further
assumptions, this also gives the data-generating value.
Assume that $\frac1{N_k}\G{k}\nto\bS_{k} \succ \bz$ and that
$\frac{N_k}{N}\to\pi_k\in(0,1)$ for each $k\in\{0,\ldots,K\}$.
We collect the TRP hyperparameters into
$\phi := (\lambda_t,\lambda_p,\bet)$.

We have that:
\begin{align}
    \T\nto
    \left(
    \sum_{k=1}^K
    \pi_k\bS_k
    \right)^{-1}
    \begin{bmatrix}
        \pi_1\bS_1
        & \ldots &
        \pi_K\bS_K
    \end{bmatrix}
    := \bar\T\,.
\end{align}
Note that this quantity does not depend on $\tau$, which is asymptotically
inert, so we will not discuss it further in this section.
However, it is nevertheless critical in practical situations with small source
datasets to avoid ill-posedness.

Since the prior moments of $\ba$ are convergent, we have that
$P(\ba,\sigma^2|\phi)\to P_{\infty}(\ba,\sigma^2|\phi)$, where, writing
$\bar\B := \E_0 - [\,\bz \;\; \bar\T\,]$ for the limit of $\B$,
\begin{align}
    P_{\infty}(\ba,\sigma^2|\phi)
    = \frac1{\sigma^2}
    \frac{\lambda_t^{P}}{Z_d}\,
    d\!\left(\lambda_t\Vert\bar\B\ba\Vert\right)
    N\!\left(\bs;\, \bz,\, \sigma^2\bL_0^{-1}\right)
    \,,
    \qquad
    Z_d := \int_{\mathbb{R}^P} d(\Vert\mathbf{u}\Vert)\,d\mathbf{u}
    \,.
    \label{eq:Pinf}
\end{align}
The middle factor is the TRP itself with $\T$ replaced by its limit, written as a normalized density on the transformed coordinate $\z=\B\ba\in\mathbb{R}^P$.

Under our model assumptions,
$P(\y_A|\ba,\sigma^2,\X_A)\to\delta_{[\ba=\bar{\be}_A,\sigma^2=\bar\sigma^2]}$.
Therefore,
\begin{align*}
    P(\phi|\y_A,\X_A)
    = \int P(\phi,\ba,\sigma^2|\y_A,\X_A) \, d\ba \, d\sigma^2
    \propto
    \int P(\y_A|\ba,\sigma^2,\X_A) \, P(\ba,\sigma^2 | \phi) \, d\ba \, d\sigma^2
    \,,
\end{align*}
such that
\begin{align*}
    & \lim_{N\to\infty} P(\phi|\y_A,\X_A) \propto
    \int \lim_{N\to\infty}
    P(\y_A|\ba,\sigma^2,\X_A) \, P(\ba,\sigma^2 | \phi) \, d\ba \, d\sigma^2
    \\ & =
    \int
    \delta_{[\ba=\bar{\be}_A,\sigma^2=\bar\sigma^2]}
    P(\ba,\sigma^2|\phi) \, d\ba \, d\sigma^2
    =
    P_\infty(\bar{\be}_A,\bar\sigma^2|\phi)\,.
\end{align*}

This gives us that:
\begin{align*}
    & \lim_{N\to\infty}
    P(\lambda_t,\lambda_p,\bet | \y_A,\X_A)
    \propto
    P(\bar{\be}_0|\bar{\be}_S,\lambda_t,\bet)
    P(\bar{\be}_S|\lambda_p)
    P(\lambda_t) P(\lambda_p)
    P(\bet)
    \\ &\propto
    \left[
    \lambda_t^P\,
    d\!\left(\lambda_t \Vert \bar\be_0 - \bar\T \bar{\be}_S\Vert\right)
    P(\lambda_t)
    \right]
    \left[
    P(\bar\be_S|\lambda_p)
    P(\lambda_p)
    \right]
    P(\bet) \,.
\end{align*}
This gives us the asymptotic posterior of the TRP hyperparameters.

Now, let us determine the marginal distribution of $\bet$.
Throughout, write
$r_{\bet} := \Vert\bar\be_0 - \bar\T_{\bet}\bar\bs\Vert$
for the limiting discrepancy between the target coefficients and the
transferred source coefficients under configuration $\bet$, measured in the
TRP's own norm.
We see that the term involving $\lambda_p$ is not involved in determining
$\bet$, so dropping it and proceeding, we obtain:
\begin{align*}
    & \lim_{N\to\infty}
    P(\bet | \y_A,\X_A) =
    \lim_{N\to\infty}
    \int P(\lambda_t,\lambda_p,\bet | \y_A,\X_A) \, d\lambda_t \, d\lambda_p
    \\ & \propto
    \left[
    \int_0^\infty
    \lambda_t^{P}\,
    d\!\left(\lambda_t r_{\bet}\right)
    P(\lambda_t)
    \, d\lambda_t
    \right]
    \times
    P(\bet) \,,
\end{align*}
where the $\lambda_t^{P}$ is the normalizing constant of the TRP.

At this point, we split into two cases, depending on whether
$r_{\bet}$ is zero or not.
First, take the case that it is zero. Then we are left with the integral:
\begin{align*}
    d(0)
    \int_0^\infty
    \lambda_t^{P}
    P(\lambda_t)
    \, d\lambda_t \,,
\end{align*}
which gives the $P$th prior moment of $\lambda_t$.
If $P(\lambda_t)$ does not have such a moment, the integral diverges, and
asymptotically, $P(\bet|\y_A,\X_A)$ concentrates entirely on this
configuration, provided it is the only configuration achieving exact transfer.

If $P(\lambda_t)$ does have this moment, then instead the integral above
evaluates to it, and the marginal posterior of $\bet$ fails to concentrate in
the limit.
This is the case, for instance, under the gamma prior on $\lambda_t^2$ that is
conventional in the Bayesian lasso \citep{park2008bayesian}, which has all
moments finite; the heavy tail of the Half Cauchy is essential here.

If $r_{\bet}$ is nonzero, we study the integral in the regime of large
$r_{\bet}$, which is the regime of interest: it is when the anticipated and
actual target coefficients are far apart that we wish to understand how sharply
the posterior penalizes a configuration.
The substitution $u = \lambda_t r_{\bet}$ moves all dependence on $r_{\bet}$
into the prior:
\begin{align*}
    \int_0^\infty
    \lambda_t^{P}\,
    d\!\left(\lambda_t r_{\bet}\right)
    P(\lambda_t)
    \, d\lambda_t
    =
    r_{\bet}^{-(P+1)}
    \int_0^\infty
    u^{P}\, d(u)\,
    P\!\left(\frac{u}{r_{\bet}}\right)
    \, du \,.
\end{align*}
Since $P(\lambda_t)$ is bounded and $u^P d(u)$ is integrable, dominated
convergence gives
\begin{align*}
    \int_0^\infty
    u^{P}\, d(u)\,
    P\!\left(\frac{u}{r_{\bet}}\right)
    \, du
    \;\xrightarrow[r_{\bet}\to\infty]{}\;
    P(0)\int_0^\infty u^{P}\, d(u)\, du
    =: C_d \,,
\end{align*}
a constant that does not depend on $\bet$.
Therefore
\begin{align*}
    \lim_{N\to\infty}
    P(\bet | \y_A,\X_A)
    \propto
    \frac{P(\bet)}{r_{\bet}^{P+1}}
    \left(1 + o(1)\right)
    \qquad\textrm{as } r_{\bet}\to\infty \,,
\end{align*}
for any prior on $\lambda_t$ satisfying our assumptions.

The order of the correction depends on the behaviour of $P(\lambda_t)$ near
zero.
To match the main article, we now take $P(\lambda_t)$ to be standard Half
Cauchy, for which
$P(u/r_{\bet}) - P(0) = -\frac{2}{\pi}\frac{(u/r_{\bet})^2}{1+(u/r_{\bet})^2}$,
so that $|P(u/r_{\bet}) - P(0)| \leq \frac{2}{\pi} u^2 r_{\bet}^{-2}$ and
\begin{align*}
    \left|
    \int_0^\infty
    u^{P}\, d(u)\,
    P\!\left(\frac{u}{r_{\bet}}\right)
    \, du
    - C_d
    \right|
    \leq
    \frac{2}{\pi\, r_{\bet}^{2}}
    \int_0^\infty u^{P+2}\, d(u)\, du \,.
\end{align*}
Hence, under the Half Cauchy prior,
\begin{align*}
    \lim_{N\to\infty}
    P(\bet | \y_A,\X_A)
    \propto
    \frac{P(\bet)}{r_{\bet}^{P+1}}
    \left(1 + O\!\left(r_{\bet}^{-2}\right)\right)
    \,.
\end{align*}
The correction is $O(r_{\bet}^{-2})$ rather than $O(r_{\bet}^{-1})$ because
$P'(0)=0$ for the Half Cauchy, as for any density symmetric about the origin
restricted to $\mathbb{R}_+$; a prior with $P'(0)\neq0$ would give
$O(r_{\bet}^{-1})$.

\section{MAP Change of Variables Details}
\label{ap:map_cov}

Starting with our MAP cost, we first define $\bt = \T\bs$, split variables,
and then proceed to form the Lagrangian associated with that constraint.
We omit the source penalty $-\log P(\bs)$ throughout this appendix; it would simply imply some penalty on the transformed variable $\bt$, and we are rather interested in refactoring the likelihood to better fit the main TRP component.
Proceeding yields:
\begin{align*}
    & \min_{\be_0,\bs} \,
    \frac1{2}\Vert \be_0 - \hbe_0\Vert_{\G{0}}^2
    +
    \frac1{2}\Vert \bs - \hat\bs\Vert_{\G{S}}^2
    +
    \lambda_t \Vert \be_0 - \T\bs \Vert_1
    \\ \iff
    & \min_{\be_0,\bs,\bt} \,
    \frac1{2}\Vert \be_0 - \hbe_0\Vert_{\G{0}}^2
    +
    \frac1{2}\Vert \bs - \hat\bs\Vert_{\G{S}}^2
    +
    \lambda_t \Vert \be_0 - \bt \Vert_1
    \\& \quad\quad \textrm{s.t.}
    \quad \bt = \T\bs
    \\ \iff &
    \min_{\be_0,\bs,\bt}
    \max_{\lagvar\in\mathbb{R}^P}
    \,
    \frac1{2}\Vert \be_0 - \hbe_0\Vert_{\G{0}}^2
    +
    \frac1{2}\Vert \bs - \hat\bs\Vert_{\G{S}}^2
    +
    \lambda_t \Vert \be_0 - \bt \Vert_1
    +
    \lagvar^\top(\bt - \T\bs)
    \,.
\end{align*}

Next, we evaluate the stationarity condition with respect to $\bs$ to obtain
the conditional minimizer $\bs^*$, implicitly a function of $\lagvar$:
\begin{align}
    & \G{S}(\bs^* - \hat\bs) - \T^\top\lagvar = \bz
    \implies
    \bs^*= \hat\bs + \Gi{S}\T^\top\lagvar \,.
    \label{eq:bss}
\end{align}

We next plug $\bs^*$ into the constraint to obtain the stationary value
$\lagvar^*$ given $\bt$:
\begin{align}
    & \bt  = \T\bs^* = \T(\hat\bs + \Gi{S}\T^\top\lagvar^*)
    =
    \hat\bt + \T\Gi{S}\T^\top \lagvar^*
    \\ & \iff
    \lagvar^* = \left(\T\Gi{S}\T^\top\right)^{-1}
    (\bt-\hat\bt)\,.
    \label{eq:muss}
\end{align}

Plugging Equation \ref{eq:muss} into Equation \ref{eq:bss} allows us to
re-express the source-data difference as follows:
\begin{align*}
    \bs^* - \hat\bs
    =
    \Gi{S}\T^\top
    \left(\T\Gi{S}\T^\top\right)^{-1}
    (\bt-\hat\bt) \,,
\end{align*}
and thus we have that:
\begin{align}
    & \Vert\bs^* - \hat\bs \Vert_{\G{S}}^2
    \\ & =
    (\bt-\hat\bt)^\top
    \left(\T\Gi{S}\T^\top\right)^{-1}
    \T\Gi{S}
    \G{S}
    \Gi{S}\T^\top
    \left(\T\Gi{S}\T^\top\right)^{-1}
    (\bt-\hat\bt)
    \\ & =
    (\bt-\hat\bt)^\top
    \left(\T\Gi{S}\T^\top\right)^{-1}
    (\bt-\hat\bt)
    = \Vert \bt-\hat\bt\Vert_{\left(\T\Gi{S}\T^\top\right)^{-1}}^2 \,.
    \label{eq:Ddef}
\end{align}

This allows us to develop the desired cost profile:
\begin{align*}
\min_{\be_0,\bt} \,
    \frac1{2}\Vert \be_0 - \hbe_0\Vert_{\TI}^2
    +
    \frac1{2}\Vert \bt-\hat\bt\Vert_{\SI}^2
    +
    \lambda_t \Vert \be_0 - \bt \Vert_1  \,,
\end{align*}
where, here and in Appendix \ref{ap:tl_asymp}, we write
\begin{align}
    \TI := \G{0}
    \quad\textrm{and}\quad
    \SI := \left(\T\Gi{S}\T^\top\right)^{-1}
\end{align}
for the two quadratic forms, the latter by reference to Equation
\ref{eq:Ddef}.

If $\tau = 0$, then $\SI$ has a simple form, since as given in Appendix
\ref{ap:tmath}, $\T = (\W^\top\G{S}\W)^{-1}\W^\top\G{S}$, and thence:
\begin{align*}
    & \SI =
    \left(
    \T \Gi{S} \T^\top
    \right)^{-1}
    =
    \left(
    (\W^\top\G{S}\W)^{-1}\W^\top\G{S}
    \Gi{S}
    \G{S}\W(\W^\top\G{S}\W)^{-1}
    \right)^{-1}
    \\ & =
    \W^\top\G{S}\W
    = \sum_{k=1}^K \G{k} \,.
\end{align*}

\section{Convergence of the MAP to the \texttt{Trans-Lasso}} \label{ap:tl_asymp}

\reptheorem{thm:tl_asymp}{
Assume the target and source design matrices are of full column rank, so that $\G{0}\succ\bz$ and $\sum_{k=1}^K\G{k}\succ\bz$, and let $\lambda_p=\tau=0$.
Then, for any $\y_0,\ldots,\y_K$,
\begin{align*}
\left\Vert \hbe_0^{\textrm{TRP}} - \hbe_0^{\textrm{TL}}\right\Vert_2
\leq
\frac{2\sqrt{P}\lambda_t}
{
\sqrt{\mineig(\G{0})}
\sqrt{\mineig\left(\sum_{k=1}^K \G{k}\right)}
}
\,.
\end{align*}
}

\begin{proof}

For brevity we below use the notation $\TI=\G{0}$ and $\SI=\sum_{k=1}^K\G{k}$.
We have established that the parts of the negative log posterior density
involving $\be_0$ may be expressed as:
\begin{align*}
    -\log P(\be_0, \bt | \y_A,\X_A) =
    \frac1{2}\Vert \be_0 - \hbe_0\Vert_{\TI}^2
    +
    \frac1{2}\Vert\bt-\hat\bt\Vert_{\SI}^2
    +
    \lambda_t \Vert \be_0 - \bt \Vert_1 \,.
\end{align*}
The MAP is defined as the component of $\be_0$ constituting the joint minimizer
of $-\log P(\be_0, \bt|\y_A,\X_A)$ with respect to both $\be_0$ and $\bt$:
\begin{align*}
    & \hbe_0^{\textrm{TRP}}
    = \argmin{\be_0} \min_{\bt}\, -\log P(\be_0, \bt | \y_A,\X_A)
    \\ & =
    \argmin{\be_0} \left(\frac1{2}\Vert \be_0 - \hbe_0\Vert_{\TI}^2
    \right)
    +
    \left(
     \min_{\bt}\,
    \frac1{2}\Vert\bt-\hat\bt\Vert_{\SI}^2
    +
    \lambda_t \Vert \be_0 - \bt \Vert_1
    \right)
    \\ &
    := \argmin{\be_0} q(\be_0) + \psi(\be_0)
    := \argmin{\be_0} \ctrp(\be_0)
    \,,
\end{align*}
where the function $q$ gives half the ($\bt$-independent) SSE on the target
dataset of a particular $\be_0$ and $\psi$ denotes the $\bt$-dependent parts of
the negative log density after minimizing over $\bt$.
We will denote the $\bt$ achieving the inner minimum as $\bt^*(\be_0)$, and
drop the dependence on $\be_0$ for notational convenience.
The cost function in its entirety is denoted $\ctrp$.

The modified Transfer Lasso is expressible as:
\begin{align*}
    & \hbe_0^{\textrm{TL}} = \argmin{\be_0}
    \frac1{2}\Vert \be_0 - \hbe_0\Vert_{\TI}^2
    + \lambda_t \Vert \be_0 - \hat\bt\Vert_1
    := \argmin{\be_0} q(\be_0) + \lambda_t\Vert \be_0 - \hat\bt\Vert_1
    := \argmin{\be_0} \ctl(\be_0)\,.
\end{align*}

Our proof proceeds in two main steps.
First, we establish a bound on $\Vert \bt^* - \hat\bt\Vert_2$.
Then, we use this to uniformly bound $\ctrp - \ctl$, which, together with our
strong convexity assumptions, leads to a bound on the desired
$\hbe_0^{\textrm{TRP}} - \hbe_0^{\textrm{TL}}$, by using the following classical
result from convex analysis:
\begin{proposition}
    \textbf{Uniform Closeness of Strongly Convex Functions Implies Closeness of
    their Optimizing Arguments} \citep[e.g.][the precise statement has been
    adapted for this article.]{hjort2011asymptotics}.
    Let $g$ and $h$ be both $m$-strongly convex, by which we mean that the maps
    $\x\to g(\x)-\frac{m}{2}\Vert \x \Vert_2^2$ and
    $\x\to h(\x)-\frac{m}{2}\Vert \x \Vert_2^2$ are convex, and denote their
    minimizers by $\x^*_g$ and $\x^*_h$.
    Assume that they are uniformly within $\epsilon$ of each other, i.e.
    $|g(\x) - h(\x)| \leq \epsilon$ for all $\x$.
    Then their minimizers are close:
    \begin{align*}
        \Vert \x^*_g - \x^*_h \Vert_2 \leq
        2\sqrt{\frac{\epsilon}{m}}\,.
    \end{align*}
    \label{prop:sc_min}
\end{proposition}

We begin our first aim of bounding the distance between $\bt^*$ and $\hat\bt$
by developing the Fermat condition associated with $\bt^*$'s optimality:
\begin{align*}
    & \bt^* \in \argmin{\bt}
    \frac1{2}\Vert\bt-\hat\bt\Vert_{\SI}^2
    +
    \lambda_t \Vert \be_0 - \bt \Vert_1
    \\ & \iff
    \bz \in \SI(\bt^* - \hat\bt) + \lambda_t \partial \Vert \be_0 - \bt^* \Vert_1
    \iff
    \bt^* - \hat\bt = \lambda_t \SI^{-1} \mathbf{s}
    \,.
\end{align*}

Here, $\mathbf{s}$ is an element of the subdifferential of the mapping
$\x\to \Vert \be_0 - \x\Vert_1$.
To evaluate this subdifferential, we again invoke a classical result in convex
analysis.
\begin{proposition}
    \textbf{Subdifferentiation under Affine Composition}
    \citep[][Theorem 23.9; modified for our setting]{rockafellar1970convex}.
    Let $f = h\circ a$ where $h$ is proper convex and
    $a(\x) = \A\x + \mathbf{b}$ is an affine map with $\A$ square and full rank.
    Then,
    \begin{align}
        \partial_{\x} f(\x) = \A^\top \, \partial h(\A\x + \mathbf{b})\,.
    \end{align}
\end{proposition}

Applied to our problem, we find that
$\partial_{\bt} \Vert \be_0 - \bt\Vert_1 = \sgn(\bt-\be_0)$.
Since $\mathbf{s} \in \sgn(\bt^* - \be_0)$, its elements are always bounded
within $[-1,1]$ regardless of $\be_0$; by standard norm bounds, this means that
$\Vert \mathbf{s}\Vert_2 \leq \sqrt{P} \Vert \mathbf{s}\Vert_\infty = \sqrt{P}$.
Therefore:
\begin{align*}
    \Vert\bt^* - \hat\bt\Vert_2 = \Vert\lambda_t \SI^{-1} \mathbf{s}\Vert_2
    \leq
    \lambda_t \frac{\Vert \mathbf{s} \Vert_2}{\mineig(\SI)}
    \leq
    \lambda_t \frac{\sqrt{P}}{\mineig(\SI)} \,,
\end{align*}
which resolves our first aim of bounding the difference between the total
source coefficient estimates used by each method.

\vspace{1em}

Our next goal is to translate this bound into a uniform one on the difference
between the TRP and TL cost functions.
Since the TL cost may be obtained by plugging in a specific value for $\bt$
while the TRP is finding the optimal such value, it must be the case that
$\ctrp(\be_0)\leq \ctl(\be_0)$ for any $\be_0$.
Therefore:
\begin{align*}
    & 0 \leq \ctl(\be_0) - \ctrp(\be_0)
    =
    \lambda_t \Vert \be_0 - \hat\bt\Vert_1
    - \psi(\be_0)
    =
    \lambda_t \Vert \be_0 - \hat\bt\Vert_1
    - \frac12\Vert \bt^* - \hat\bt\Vert_{\SI}^2
    - \lambda_t \Vert \be_0 - \bt^*\Vert_1
    \\ &
    \leq
    \lambda_t \left(\Vert \be_0 - \hat\bt\Vert_1
    -  \Vert \be_0 - \bt^*\Vert_1 \right)
    \leq
    \lambda_t \Vert \bt^* - \hat\bt\Vert_1
    \leq \lambda_t \sqrt{P} \Vert \bt^* - \hat\bt\Vert_2 \leq
    \frac{\lambda^2_t P }{\mineig(\SI)}
    := \epsilon \,.
\end{align*}
This gives us therefore that
$\Vert \ctl - \ctrp\Vert_{\infty}\leq \epsilon$.

\vspace{1em}

Since both costs have a strong convexity constant given by $\mineig(\TI)$, our
final result follows from applying Proposition \ref{prop:sc_min}:
\begin{align}
    \Vert\hbe_0^{\textrm{TRP}} - \hbe_0^{\textrm{TL}}\Vert_2 \leq
    2\sqrt{\frac{\epsilon}{\mineig(\TI)}} =
    \frac{2 \sqrt{P} \lambda_t}{\sqrt{\mineig(\SI)\mineig(\TI)}} \,.
\end{align}

\end{proof}

\section{Derivation of the Proximal Operators}
\label{ap:prox}

In this section, we derive the solutions of the TRP-MAP estimator in the
univariate / sample means / orthonormal setting.
Throughout, $\bar{y}_0$ and $\bar{y}_1$ denote the target and source sample
means, $s_0$ and $s_1$ their sampling variances, and $\mu_0,\mu_1$ the
corresponding optimization variables, as in Section \ref{sec:univariate}.

\subsection{Derivation of the $\lambda_p=0$ prox}
\label{ap:lamp0}

Though implied by our subsequent more general result, we nevertheless here
study the problem with no source penalty as its proof is much simpler than the
general case, that is, we now study the problem:
\begin{align*}
    \min_{\mu_0,\mu_1} \sr{\mu_0-\bar{y}_0}{s_0} + \sr{\mu_1-\bar{y}_1}{s_1}
    + \lambda_t|\mu_0-\mu_1| \,.
\end{align*}

The transformation $w = \frac{s_1 \mu_0 + s_0 \mu_1}{s_0+s_1}$,
$\delta = \mu_0-\mu_1$ yields inverse transforms:
\begin{align*}
    \mu_1 = w - \kappa \delta \,,\quad
    \mu_0 = w + (1-\kappa) \delta \,,
\end{align*}
with $\kappa = \frac{s_1}{s_0+s_1}$.
This yields the problem:
\begin{align*}
    \min_{w,\delta}
    \sr{w+(1-\kappa)\delta - \bar{y}_0}{s_0}
    +
    \sr{w-\kappa \delta - \bar{y}_1}{s_1}
    + \lambda_t|\delta|
    :=
    \min_{w,\delta}
    l(w,\delta)
    \,.
\end{align*}

We can directly inspect the first-order conditions, which will show that the
two variables decouple since $\frac{1-\kappa}{s_0} = \frac{\kappa}{s_1}$:
\begin{align*}
    &0 = \frac{\partial l(w,\delta)}{\partial w}
    =
    \frac{w - \bar{y}_0}{s_0}
    +
    \frac{w-\bar{y}_1}{s_1}
    \iff
    \boxed{w^* = \frac{s_1 \bar{y}_0 + s_0 \bar{y}_1}{s_0+s_1}
    := \bar{w}}\,;
    \\
    &
    0 \in \partial_{\delta} l(w,\delta) =
    \frac{1-\kappa}{s_0}
    ((1-\kappa)\delta + w - \bar{y}_0)
    -
    \frac{\kappa}{s_1}
    (-\kappa \delta + w - \bar{y}_1)
    + \lambda_t \partial|\delta|
    \\ & \iff
    0 \in
    \frac1{s_0+s_1}
    \left(
    \delta - (\bar{y}_0-\bar{y}_1)
    \right)
    + \lambda_t \partial|\delta|
    \iff
    \bar{y}_0-\bar{y}_1 - \delta
    \in
    (s_0+s_1)\lambda_t \, \partial |\delta|
    \\ & \iff
    \boxed{\delta^* = S(\bar{y}_0-\bar{y}_1, (s_0+s_1)\lambda_t)}
    \,.
\end{align*}

Subsequently,
\begin{align*}
    & \mu_0^* = \bar{w} + \frac{s_0}{s_0+s_1} S(\bar{y}_0-\bar{y}_1,(s_0+s_1)\lambda_t) \\
    & \mu_1^* = \bar{w} - \frac{s_1}{s_0+s_1} S(\bar{y}_0-\bar{y}_1,(s_0+s_1)\lambda_t) \,.
\end{align*}

\subsection{Derivation of the $\lambda_p>0$ prox}
\label{ap:lamp_big_prox}

In this subsection we study the problem:
\begin{align*}
    \min_{\mu_0,\mu_1}
    \sr{\mu_0-\bar{y}_0}{s_0}+
    \sr{\mu_1-\bar{y}_1}{s_1}+
    \lambda_t|\mu_0-\mu_1|
    + \lambda_p|\mu_1| \,.
\end{align*}

We will show that the solution is given by:
\begin{align*}
    &
    \zeta \gets S\left(\frac{s_1 \bar{y}_0 + s_0 \bar{y}_1}{s_0+s_1}, \frac{\lambda_p}{\frac{1}{s_0}+\frac{1}{s_1}}\right)
    \\ &
    \mu_0^* \gets \zeta + S\left(\bar{y}_0 - \zeta, s_0\lambda_t\right)\,.
\end{align*}

We proceed by splitting variables and dualizing, introducing the scalar
multiplier $\nu$:
\begin{align*}
    & \min_{\mu_0,\mu_1}
    \sr{\mu_0-\bar{y}_0}{s_0}+
    \sr{\mu_1-\bar{y}_1}{s_1}+
    \lambda_t|\mu_0-\mu_1|
    + \lambda_p|\mu_1|
    \\ & \iff
    \min_{\mu_0,\mu_1,\delta}
    \sr{\mu_0-\bar{y}_0}{s_0}+
    \sr{\mu_1-\bar{y}_1}{s_1}
    +\lambda_t|\delta|
    +\lambda_p|\mu_1|
    + \max_{\nu} \nu(\delta-\mu_0+\mu_1)
    \\ &\iff
    \max_{\nu}
    \,
    C(\nu)+
    \left[
    \min_{\mu_0}
    \sr{\mu_0-(\bar{y}_0+s_0\nu)}{s_0}
    \right]
    +
    \left[
    \min_{\mu_1}
    \sr{\mu_1-(\bar{y}_1-s_1\nu)}{s_1}
    +\lambda_p|\mu_1|
    \right]
    +
    \left[
    \min_{\delta}
    \lambda_t |\delta| + \nu \delta
    \right] \,,
\end{align*}
where the first term $C(\nu)$ does not depend on $\mu_0,\mu_1,\delta$ and is
given explicitly as
$-\frac{s_0+s_1}{2}\left(\nu-\frac{\bar{y}_1-\bar{y}_0}{s_0+s_1}\right)^2$.

The variables $\mu_0,\mu_1$ have clear solutions:
\begin{align*}
    & \mu_0^* = \bar{y}_0 + s_0\nu
    \\ &
    \mu_1^* = S\left(
    \bar{y}_1 - s_1\nu, s_1 \lambda_p
    \right) \,.
\end{align*}
The $\delta$ optimization actually imposes the constraint that
$|\nu|\leq\lambda_t$, in which case $\delta^* = 0$.

To determine the profile objective for $\nu$, we plug in those optima and
augment the $C(\nu)$ term with them.

\textbf{Contribution of $\mu_0$.}
Since the $\mu_0$ minimization is just a quadratic, it has a direct
contribution of $0$ to the objective beyond $C(\nu)$.

\textbf{Contribution of $\mu_1$.}
We have:
\begin{align*}
    & \mu_1^* = S(\bar{y}_1 - s_1\nu, s_1\lambda_p)\,;
    \qquad
    \sr{\mu_1^* - (\bar{y}_1-s_1\nu)}{s_1}
    + \lambda_p|\mu_1^*| \,.
\end{align*}
As we will see, this contribution is the Huber loss in $\nu$.
Worth noting first is that:
\begin{align*}
    &
    S(\bar{y}_1-s_1\nu, \lambda_p s_1) =
    \sgn(\bar{y}_1-s_1\nu)
    \left(|\bar{y}_1-s_1\nu| - \lambda_p s_1\right)^+
    \\ & =
    - s_1\sgn\left(\nu-\frac{\bar{y}_1}{s_1}\right)
    \left(\left|\nu - \frac{\bar{y}_1}{s_1}\right| - \lambda_p \right)^+
    = -s_1 S\!\left(\nu-\frac{\bar{y}_1}{s_1},\lambda_p\right)\,.
\end{align*}
Therefore,
\begin{align*}
    \lambda_p|\mu_1^*| =
    s_1\lambda_p\left|S\!\left(\nu-\frac{\bar{y}_1}{s_1},\lambda_p\right)\right|
    = s_1\lambda_p\left(\left|\nu-\frac{\bar{y}_1}{s_1}\right|-\lambda_p\right)^+\,.
\end{align*}

As to the other term:
\begin{align*}
    &
    (\mu_1^* - (\bar{y}_1-s_1\nu))^2 =
    (S(\bar{y}_1-s_1\nu,s_1\lambda_p) - (\bar{y}_1-s_1\nu))^2
    \\ & =
    \left(
    \sgn(\bar{y}_1-s_1\nu)\left[
    \left(
    |\bar{y}_1-s_1\nu| - s_1\lambda_p
    \right)^+
    -|\bar{y}_1-s_1\nu|
    \right]
    \right)^2
    \\ & =
    \left(
    \max\left(0,
    |\bar{y}_1-s_1\nu| - s_1\lambda_p
    \right)
    -|\bar{y}_1-s_1\nu|
    \right)^2
    \\ & =
    \left(
    \max\left(-|\bar{y}_1-s_1\nu|,
    - s_1\lambda_p
    \right)
    \right)^2
    =
    \min\!\left((\bar{y}_1-s_1\nu)^2, s_1^2\lambda_p^2\right)
    \\ & = s_1^2 \min\!\left(\left(\nu - \frac{\bar{y}_1}{s_1}\right)^2, \lambda_p^2\right)\,.
\end{align*}

\textbf{Taken together}, we therefore have the following $\mu_1$ contribution:
\begin{align*}
    s_1 \left[
    \frac{\min\left(\left(\nu - \frac{\bar{y}_1}{s_1}\right)^2,\lambda_p^2\right)}{2}
    + \lambda_p \left(\left|\nu - \frac{\bar{y}_1}{s_1}\right| - \lambda_p\right)^+
    \right]\,.
\end{align*}

By considering separately the cases where $\left|\nu-\frac{\bar{y}_1}{s_1}\right|$
is above or below $\lambda_p$, we find that this is expressible as:
\begin{align*}
    s_1
    \begin{cases}
        \frac12\left(\nu-\frac{\bar{y}_1}{s_1}\right)^2 & \left|\nu - \frac{\bar{y}_1}{s_1}\right|\leq \lambda_p \\
        \lambda_p \left|\nu - \frac{\bar{y}_1}{s_1}\right| -\frac{\lambda_p^2}{2} & \textrm{o.w.}
    \end{cases}\,,
\end{align*}
which is recognized as the Huber loss function.
It is probably worth re-emphasizing that we are currently \textit{maximizing}
with respect to $\nu$, whereas the Huber \textit{loss} is typically found in
minimization problems.
Despite its piecewise definition, the Huber loss is continuously differentiable
everywhere.
Let us now bring back in the concave quadratic term to form the complete
optimization problem:
\begin{align}
    & \max_{\nu\in[-\lambda_t,\lambda_t]}\,
    -\frac{s_0+s_1}{2}\left(\nu-\frac{\bar{y}_1-\bar{y}_0}{s_0+s_1}\right)^2
    +
    s_1
    \begin{cases}
        \frac12\left(\nu-\frac{\bar{y}_1}{s_1}\right)^2 & \left|\nu - \frac{\bar{y}_1}{s_1}\right|\leq \lambda_p \\
        \lambda_p \left|\nu - \frac{\bar{y}_1}{s_1}\right| -\frac{\lambda_p^2}{2} & \textrm{o.w.}
    \end{cases}
    \\ & =
    \label{eq:expanded}
    \max_{\nu\in[-\lambda_t,\lambda_t]}\,
    -\frac{s_0}{2}\left(\nu + \frac{\bar{y}_0}{s_0}\right)^2
    + s_1
    \begin{cases}
        0 & \left|\nu - \frac{\bar{y}_1}{s_1}\right|\leq \lambda_p \\
        -\frac12 \left(\nu-\frac{\bar{y}_1}{s_1}\right)^2 + \lambda_p \left|\nu - \frac{\bar{y}_1}{s_1}\right| -\frac{\lambda_p^2}{2} & \textrm{o.w.}
    \end{cases}
    \,.
\end{align}

We notice that the convex quadratic from the $\mu_1$ contribution has
coefficient $s_1$ while the concave quadratic has coefficient $s_0+s_1$, such
that the overall objective is still concave as expected by the basic theory of
Lagrangian duality.
Optimization with respect to $\nu$ may be achieved simply by searching for a
stationary point of this function.

The first question is: when do we expect to have
$\left|\nu^* - \frac{\bar{y}_1}{s_1}\right|\leq \lambda_p$?
In this case, $\nu^*$ is simply the solution to the first term, given by
$-\frac{\bar{y}_0}{s_0}$, such that the condition reads:
\begin{align*}
    & \left|-\frac{\bar{y}_0}{s_0}-\frac{\bar{y}_1}{s_1}\right| \leq \lambda_p
    \iff \left|\frac{\bar{y}_0}{s_0} + \frac{\bar{y}_1}{s_1}\right|\leq \lambda_p
    \\ &  \iff
    \left|
    \frac{\frac{\bar{y}_0}{s_0}+\frac{\bar{y}_1}{s_1}}{\frac1{s_0}+\frac1{s_1}}
    \right|
    \leq
    \frac{\lambda_p}
    {\frac1{s_0}+\frac1{s_1}}
    \iff |\bar{w}| \leq
    \frac{\lambda_p}
    {\frac1{s_0}+\frac1{s_1}}
    := \tilde{\lambda}_p \,.
\end{align*}

If not, that is, if $|\bar{w}| > \tilde\lambda_p$, then $\nu^*$ solves the
stationarity condition for the second branch of Equation \ref{eq:expanded},
\begin{align*}
    -s_0\left(\nu^* + \frac{\bar{y}_0}{s_0}\right)
    + s_1\left[
    -\left(\nu^*-\frac{\bar{y}_1}{s_1}\right)
    + \lambda_p \sgn\!\left(\nu^* - \frac{\bar{y}_1}{s_1}\right)
    \right] = 0 \,,
\end{align*}
which rearranges to
\begin{equation}
    \nu^* = \frac{\bar{y}_1-\bar{y}_0}{s_0+s_1}
    + \frac{s_1\lambda_p}{s_0+s_1} \sgn\!\left(\nu^* - \frac{\bar{y}_1}{s_1}\right) \,.
\end{equation}
Now we must determine $\sgn\left(\nu^*-\frac{\bar{y}_1}{s_1}\right)$.
Looking at Equation \ref{eq:expanded}, we notice that $\nu^*$ must lie between
$\frac{-\bar{y}_0}{s_0}$ and $\frac{\bar{y}_1}{s_1}$.
Therefore,
$\sgn\left(\nu^* - \frac{\bar{y}_1}{s_1}\right) = \sgn\left(-\frac{\bar{y}_0}{s_0}-\frac{\bar{y}_1}{s_1}\right) = -\sgn(\bar{w})$.
Thus, we have:
\begin{equation}
    \nu^* = \frac{\bar{y}_1-\bar{y}_0}{s_0+s_1}
    - \frac{s_1}{s_0+s_1} \lambda_p\sgn(\bar{w}) \,.
\end{equation}

Plugging these two cases in yields:
\begin{align*}
    & \nu^* \gets \begin{cases}
        -\frac{\bar{y}_0}{s_0} & |\bar{w}|\leq \tilde\lambda_p \\
        -\frac{s_1\lambda_p\sgn(\bar{w})}{s_0+s_1} - \frac{\bar{y}_0-\bar{y}_1}{s_0+s_1} & \textrm{o.w.} \\
    \end{cases} \\
    & \tilde{\nu} \gets \textrm{Clip}(\nu^*; -\lambda_t,\lambda_t) \\
    & \mu_0^* \gets \bar{y}_0 + s_0\tilde\nu  \,.
\end{align*}

First, we notice that:
\begin{align}
    & \zeta := \bar{y}_0  + s_0 \nu^*
    =
    \begin{cases}
        \bar{y}_0 - s_0 \left(\frac{\bar{y}_0}{s_0}\right) & |\bar{w}| \leq \tilde\lambda_p \\
        \bar{y}_0 - \frac{s_0 s_1}{s_0+s_1} \lambda_p \sgn(\bar{w}) - \frac{s_0}{s_0+s_1}(\bar{y}_0 - \bar{y}_1) & \textrm{o.w.}
    \end{cases}
    \\
    & =
    \begin{cases}
        0 & |\bar{w}| \leq \tilde\lambda_p \\
        \bar{w} - \tilde\lambda_p \sgn(\bar{w}) & \textrm{o.w.}
    \end{cases}
    =
    S(\bar{w}, \tilde\lambda_p)\,.
\end{align}

So this gives us $\bar{y}_0 + s_0\nu^*$.
But actually, what we want is
$\bar{y}_0 + s_0 \, \textrm{Clip}(\nu^*;-\lambda_t,\lambda_t)$.
Since $\textrm{Clip}(\nu;-\lambda_t,\lambda_t) = \nu - S(\nu, \lambda_t)$, we
have that:
\begin{align}
    & \mu_0^* = \bar{y}_0 + s_0 \, \textrm{Clip}(\nu^*;-\lambda_t,\lambda_t)
    =
    \bar{y}_0 + s_0(\nu^* - S(\nu^*,\lambda_t))
    =
    \bar{y}_0 + s_0(\nu^* + S(-\nu^*,\lambda_t))
    \\
    & =
    \bar{y}_0 + s_0\nu^* + S(-s_0\nu^*,s_0\lambda_t)
    =
    \bar{y}_0 + s_0\nu^* + S\!\left(\bar{y}_0 - (\bar{y}_0 +s_0\nu^*),\,s_0\lambda_t\right)
    = \zeta + S(\bar{y}_0 - \zeta,s_0\lambda_t) \,,
\end{align}
as desired.

\section{Normal Means Risk Comparison}
\label{ap:risk}

Let $S(x,t)$ be the soft thresholding operator and $\delta = \bar{y}_0-\bar{y}_1$,
so that $\delta \sim N(\mu_0-\mu_1, s_0+s_1)$.
Our two estimators are given by:
\begin{align}
    & \hat{\mu}_0^{\textrm{TL}} = \bar{y}_1 + S(\delta,\lambda_t s_0) \\
    & \hat{\mu}_0^{\textrm{TRP}} =
    \frac{\frac{\bar{y}_0}{s_0}+\frac{\bar{y}_1}{s_1}}{\frac1{s_0}+\frac1{s_1}}
    + \frac{s_0}{s_0+s_1} S\left(\delta, \lambda_t (s_0+s_1)\right)\,.
\end{align}

Let $c=\frac{s_0}{s_0+s_1}$ and $m = \mu_0-\mu_1$, so that $1-c = \frac{s_1}{s_0+s_1}$ and $c(s_0+s_1) = s_0$.
Then we can rewrite as:
\begin{align}
    & \hat{\mu}_0^{\textrm{TRP}} =
    (1-c)\bar{y}_0+c\,\bar{y}_1
    + c \, S\left(\delta, \lambda_t (s_0+s_1)\right) \,.
\end{align}

Now denote $\bar{w} = (1-c)\bar{y}_0+c\,\bar{y}_1$ and similarly
$\mu_w = (1-c)\mu_0 + c\,\mu_1$.
It is easy to see that $\bar{w}$ and $\delta$ are independent and that
$\mathbb{E}[\bar{w}]=\mu_w$.
We can also rewrite:
\begin{align}
    &\hat{\mu}_0^{\textrm{TL}} =
    \bar{y}_1 + S(\delta,\lambda_t s_0)
    =
    \bar{y}_1 + (1-c)(\bar{y}_0-\bar{y}_1) - (1-c) \delta + S(\delta,\lambda_t s_0)
    =
    \bar{w} + S(\delta,\lambda_t s_0) - (1-c) \delta \,.
\end{align}
Denoting $u_{\textrm{TL}}(\delta) = S(\delta,\lambda_t s_0) - (1-c)\delta$, we
find that $\hat{\mu}_0^{\textrm{TL}}$ is expressible as
$\bar{w} + u_{\textrm{TL}}(\delta)$ where the two components are independent.
This means that we can simplify the expression for MSE for
$\hat{\mu}_0^{\textrm{TL}}$ as such:
\begin{align*}
    \mathbb{E}\left[\left(
    \bar{w} + u_{\textrm{TL}}(\delta) - \mu_0
    \right)^2\right]
    =
    \mathbb{E}\left[\left(
    (\bar{w} - \mu_w)  + (u_{\textrm{TL}}(\delta) - (\mu_0-\mu_w))
    \right)^2\right]
    =
    \mathbb{V}[\bar{w}]
    +
    \mathbb{E}\left[\left(
    u_{\textrm{TL}}(\delta) - cm
    \right)^2\right]
    \,.
\end{align*}
Similarly,
$\hat{\mu}_0^{\textrm{TRP}} = \bar{w} + c \, S(\delta, \lambda_t(s_0+s_1))$, so
it has the decomposition $\bar{w} + u_{\textrm{TRP}}(\delta)$ and a similar MSE
decomposition.
This means we can simplify the difference in MSEs as such:
\begin{align*}
    \mathbb{E}\left[\left(
    \hat{\mu}_0^{\textrm{TL}} - \mu_0
    \right)^2\right]
    -
    \mathbb{E}\left[\left(
    \hat{\mu}_0^{\textrm{TRP}} - \mu_0
    \right)^2\right]
    =
    \mathbb{E}\left[\left(
    u_{\textrm{TL}}(\delta) - mc
    \right)^2\right]
    -
    \mathbb{E}\left[\left(
    u_{\textrm{TRP}}(\delta) - mc
    \right)^2\right]
    := D
    \,.
\end{align*}

We wish to show that $D$ is nonnegative for any choice of
$\mu_0,\mu_1,s_0,s_1,\lambda_t$.
Note that $D=0$ when $\lambda_t=0$, since at this point
$\hat\mu_0^{\textrm{TL}}=\hat\mu_0^{\textrm{TRP}} = \bar{y}_0$.
We can thus accomplish this by showing that the partial derivative of $D$ with
respect to $\lambda_t$ is nonnegative everywhere in $\mu_0,\mu_1,s_0,s_1$.

To this end, we develop a simplified expression for the derivative by
speculatively interchanging the differentiation and integration:
\begin{align*}
    & \partial_{\lambda_t} D =
    \partial_{\lambda_t}
    \left(
    \mathbb{E}\left[\left(
    u_{\textrm{TL}}(\delta) - mc
    \right)^2\right]
    -
    \mathbb{E}\left[\left(
    u_{\textrm{TRP}}(\delta) - mc
    \right)^2\right]
    \right)
    \\ & =
    2\,\mathbb{E}\left[
    \left(
    u_{\textrm{TL}}(\delta) - mc
    \right)
    \partial_{\lambda_t} u_{\textrm{TL}}(\delta)
    -
    \left(
    u_{\textrm{TRP}}(\delta) - mc
    \right)
    \partial_{\lambda_t} u_{\textrm{TRP}}(\delta)
    \right]  \,.
\end{align*}

Towards simplifying this expression, we calculate that, almost everywhere:
\begin{align*}
    & \partial_{\lambda_t} u_{\textrm{TL}}(\delta) = -s_0\sgn(\delta)\bm{1}[|\delta|\geq \lambda_t s_0]
    \\ &
    \partial_{\lambda_t} u_{\textrm{TRP}}(\delta) = -s_0\sgn(\delta)\bm{1}[|\delta|\geq \lambda_t (s_0+s_1)]
    \,.
\end{align*}
We see now that the integrand was almost everywhere continuously
differentiable and that this interchange of limits was legitimate.
Plugging those in yields:
\begin{align*}
    & \partial_{\lambda_t} D =
    -2s_0
    \mathbb{E}\left[
    \left(
    u_{\textrm{TL}}(\delta) - mc
    \right)
    \sgn(\delta)\bm{1}[|\delta|\geq \lambda_t s_0]
    -
    \left(
    u_{\textrm{TRP}}(\delta) - mc
    \right)
    \sgn(\delta)\bm{1}[|\delta|\geq \lambda_t (s_0+s_1)]
    \right]
    \\ & =
    -2s_0
    \left(
    \mathbb{E}\left[
    \left(
    u_{\textrm{TL}}(\delta) - mc
    \right)
    \sgn(\delta)\bm{1}\!\left[|\delta| \in \lambda_t[s_0,s_0+s_1]\right]
    \right]
    +
    \mathbb{E}\left[
    \left(
    u_{\textrm{TL}}(\delta) -  u_{\textrm{TRP}}(\delta)
    \right)
    \sgn(\delta)\bm{1}[|\delta|\geq \lambda_t (s_0+s_1)]
    \right]
    \right)  \,.
\end{align*}

But we know that $\hat\mu_0^{\textrm{TL}} = \hat\mu_0^{\textrm{TRP}}$ when
$|\delta|\geq \lambda_t(s_0+s_1)$, which implies that
$u_{\textrm{TL}}(\delta)=u_{\textrm{TRP}}(\delta)$ as well, so the second term
disappears, leaving us with:
\begin{align*}
    \partial_{\lambda_t} D =
    2s_0\,
    \mathbb{E}\left[
    \left(
    mc - u_{\textrm{TL}}(\delta)
    \right)
    \sgn(\delta)\bm{1}\!\left[|\delta| \in \lambda_t[s_0,s_0+s_1]\right]
    \right]
    \,.
\end{align*}

On the interval $\lambda_t [s_0, s_0+s_1]$,
$u_{\textrm{TL}}(\delta) = \sgn(\delta)(|\delta|-\lambda_t s_0) - (1-c) \delta = c\delta - \sgn(\delta)\lambda_t s_0$,
so we plug this in:
\begin{align*}
    & \partial_{\lambda_t} D =
    2s_0\,
    \mathbb{E}\left[
    \left(
    mc - c\delta + \sgn(\delta) \lambda_t s_0
    \right)
    \sgn(\delta)\bm{1}\!\left[|\delta| \in \lambda_t[s_0,s_0+s_1]\right]
    \right]
    \\ & =
    2s_0\,
    \mathbb{E}\left[
    \left(
    \sgn(\delta) mc - c|\delta| + \lambda_t s_0
    \right)
    \bm{1}\!\left[|\delta| \in \lambda_t[s_0,s_0+s_1]\right]
    \right]
    \\ & =
    2s_0\,
    \mathbb{E}\left[
    \sgn(\delta) mc \,
    \bm{1}\!\left[|\delta| \in \lambda_t[s_0,s_0+s_1]\right]
    \right]
    +
    2s_0\,
    \mathbb{E}\left[
    \left(
    \lambda_t s_0 - c|\delta|
    \right)
    \bm{1}\!\left[|\delta| \in \lambda_t[s_0,s_0+s_1]\right]
    \right]
    \,.
\end{align*}

Start with the second expression. 
Since $|\delta| \in \lambda_t [s_0,s_0+s_1]$ and $c = \frac{s_0}{s_0+s_1}$, we have $c|\delta| \leq c\lambda_t(s_0+s_1) = \lambda_t s_0$, so that
$\lambda_t s_0 - c|\delta| \geq 0$, with equality only at the upper end of the interval, and elsewhere it is positive. 
Since $s_1>0$, this second expression is strictly positive.
On the other hand,
\begin{align*}
    &\mathbb{E}\left[
    \sgn(\delta) mc \,
    \bm{1}\!\left[|\delta| \in \lambda_t[s_0,s_0+s_1]\right]
    \right]
    \\ & =
    mc \,
    \mathbb{E}\left[
    \sgn(\delta)\,\bm{1}\!\left[|\delta| \in \lambda_t[s_0,s_0+s_1]\right]
    \right]
    = mc\left(
    P\!\left(\delta \in \lambda_t [s_0,s_0+s_1]\right)
    -
    P\!\left(\delta\in \lambda_t[-s_0-s_1,-s_0]\right)
    \right) \,,
\end{align*}
and it is clear that the sign of the probability difference matches the sign of
$m$, such that this expression is nonnegative.
Therefore, $\partial_{\lambda_t} D > 0$ everywhere, and since
\begin{align}
    D(\lambda_t) = \int_{0}^{\lambda_t} \partial_t D(t) \, dt > 0\,,
\end{align}
we have completed the proof.

\end{document}